\documentclass{article}
\usepackage[final, nonatbib]{corl_2025}

\usepackage{color}
\usepackage[usenames,dvipsnames,table]{xcolor}
\usepackage{times}
\usepackage{graphics} % for pdf, bitmapped graphics files
\usepackage{epsfig} % for postscript graphics files
\usepackage{mathptmx} % assumes new font selection scheme installed
\usepackage{times} % assumes new font selection scheme installed
\usepackage{amsmath} % assumes amsmath package installed
\usepackage{amssymb}  % assumes amsmath package installed
\usepackage{multirow}
\usepackage{multicol}
\usepackage{enumerate}
\usepackage{tabularx}
\usepackage{hyperref}
\usepackage{pifont}
\usepackage{booktabs}
\usepackage{algorithm}
\usepackage{amsmath}
\usepackage{adjustbox}
\usepackage{wrapfig}
\usepackage{anyfontsize}
\usepackage[most]{tcolorbox}
\usepackage{lipsum} % for placeholder text
\hypersetup{
    colorlinks=true,
    linkcolor=MidnightBlue,
    filecolor=magenta,      
    urlcolor=MidnightBlue,
    citecolor=MidnightBlue,
} 
\usepackage[font=small,labelfont=bf]{caption}
\usepackage[sorting=nyt,sortcites=true]{biblatex}

\title{\textbf{\textit{AirExo}-2}: Scaling up Generalizable Robotic Imitation Learning with Low-Cost Exoskeletons}

\author{
  Hongjie Fang$^{*,1}$, Chenxi Wang$^{*,2}$, Yiming Wang$^{*,1}$, Jingjing Chen$^{*,1}$, \\ \textbf{Shangning Xia}$^{1,2}$\textbf{,} \textbf{Jun Lv}$^{1,2}$\textbf{,} \textbf{Zihao He}$^1$\textbf{,} \textbf{Xiyan Yi}$^1$\textbf{,} \textbf{Yunhan Guo}$^1$\textbf{,} \textbf{Xinyu Zhan}$^1$\textbf{,}\\
  \textbf{Lixin Yang}$^1$\textbf{,} \textbf{Weiming Wang}$^1$\textbf{,} \textbf{Cewu Lu}$^{1,2,3,\dagger}$\textbf{,} \textbf{Hao-Shu Fang}$^{1,\dagger}$ \\
  $^1$Shanghai Jiao Tong University \quad $^2$Noematrix \quad
  $^3$Shanghai Innovation Institute \\
  $^*$Equal Contribution \quad $^\dagger$Corresponding Authors\\ \\
  \href{http://airexo.tech/airexo2}{\texttt{http://airexo.tech/airexo2}}
}

\begin{document}

\makeatletter
\let\@oldmaketitle\@maketitle% Store \@maketitle
\renewcommand{\@maketitle}{\@oldmaketitle% Update \@maketitle to insert...
\vspace{-0.6cm}
\centering
\includegraphics[width=0.8\linewidth]{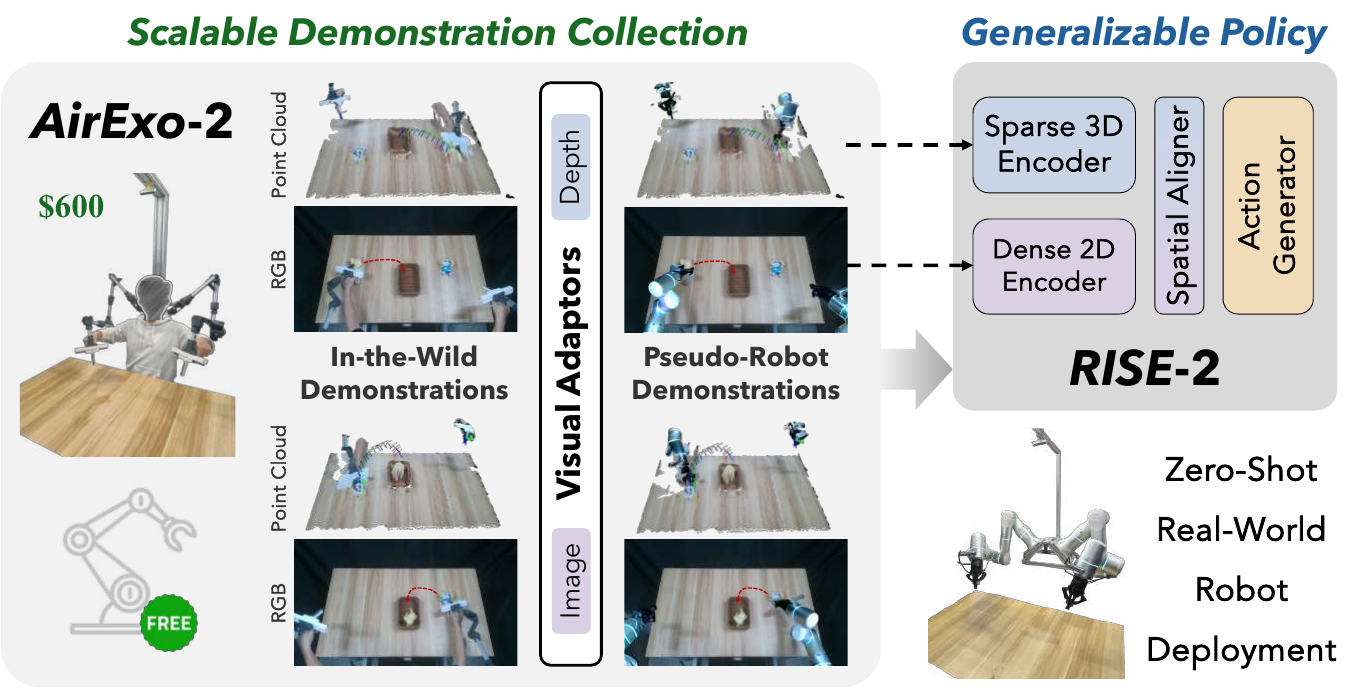}
\vspace{-0.1cm}
\captionof{figure}{\textbf{Overview of the \textit{AirExo}-2 System and the \textit{RISE}-2 Policy}. \textbf{\textit{AirExo}-2} enables scalable in-the-wild demonstration collection and adaptation. Visual adaptors convert these into pseudo-robot demonstrations for policy training. The generalizable \textbf{\textit{RISE}-2} policy learns from this data alone and achieves zero-shot deployment on real dual-arm robots, matching the performance of policies trained on teleoperated robot data.}
\label{fig:teaser}
}%
\makeatother

\maketitle

%===============================================================================

\begin{abstract}
Scaling up robotic imitation learning for real-world applications requires efficient and scalable demonstration collection methods. While teleoperation is effective, it depends on costly and inflexible robot platforms. In-the-wild demonstrations offer a promising alternative, but existing collection devices have key limitations: handheld setups offer limited observational coverage, and whole-body systems often require fine-tuning with robot data due to domain gaps. To address these challenges, we present \textbf{\textit{AirExo}-2}, a low-cost exoskeleton system for large-scale in-the-wild data collection, along with several adaptors that transform collected data into pseudo-robot demonstrations suitable for policy learning. We further introduce \textbf{\textit{RISE}-2}, a generalizable imitation learning policy that fuses 3D spatial and 2D semantic perception for robust manipulations. Experiments show that \textbf{\textit{RISE}-2} outperforms prior state-of-the-art methods on both in-domain and generalization evaluations. Trained solely on adapted in-the-wild data produced by \textbf{\textit{AirExo}-2}, the \textbf{\textit{RISE}-2} policy achieves comparable performance to the policy trained with teleoperated data, highlighting the effectiveness and potential of \textbf{\textit{AirExo}-2} for scalable and generalizable imitation learning. 
\end{abstract}


%===============================================================================

\begingroup
\renewcommand{\thefootnote}{} % Remove number
\footnotetext{$^*$ galaxies@sjtu.edu.cn, cxwang85@gmail.com, sommerfeld@sjtu.edu.cn, jjchen20@sjtu.edu.cn}
\footnotetext{$^\dagger$ lucewu@sjtu.edu.cn, fhaoshu@gmail.com}
\addtocounter{footnote}{-2} % Reset footnote counter
\endgroup

\section{Introduction}\label{sec:intro}

Scaling up generalizable robotic imitation learning in the real world is key to developing robust policies for practical applications~\cite{position_realworld, pi0}. While teleoperation is widely used for collecting demonstrations, it requires physical robots to record both observations and actions for \textit{robot-centric} demonstrations, making it costly and difficult to scale. To address this, recent work explores alternative, low-cost data sources such as human videos~\cite{gen2act, phantom, avid, instant_policy, mimicplay, atm, learning_by_watching} and in-the-wild demonstrations~\cite{umi, airexo_v1, dobbe_stickv1, dexcap, demoat}. These \textit{human-centric} demonstrations eliminate the need for a robot during collection, reducing costs and improving scalability. Human video approaches typically infer actions via hand or object pose estimation~\cite{ditto, r+x, instant_policy, im2flow2act}, while in-the-wild demonstrations use devices that interface human motion with robot-relevant data during task execution.

Current in-the-wild demonstration devices fall into two main categories: handheld~\cite{umi, rum_stickv2, legato, dobbe_stickv1, demoat} and whole-body devices~\cite{arcap, airexo_v1, m2r, dexcap}. Handheld devices are typically designed with end-effectors identical to those of the robot. By equipping both the handheld devices and the robot with in-hand cameras, these methods leverage in-hand observations to ensure visual consistency. However, they face two key issues: (1) reliance on visual SLAM for pose estimation leads to action inaccuracies (\S\ref{sec:analysis}); and (2) limited camera field of view hinders depth perception during interactions (Appendix \ref{app:case-study}). On the contrary, whole-body devices offer more accurate action capture and flexible observations, but they typically require fine-tuning with teleoperated data~\cite{airexo_v1, dexcap}, as in-the-wild demonstrations still exhibit a domain gap compared to real robot data.

To address these challenges, we present the \textbf{\textit{AirExo}-2} system for large-scale in-the-wild demonstration collection and adaptation. As illustrated in Fig.~\ref{fig:teaser} (left), we introduce demonstration adaptors that align both observations and actions from in-the-wild demonstrations with the robot domain, enabling direct use in imitation learning. On the hardware side, we build upon AirExo~\cite{airexo_v1} with a redesigned exoskeleton featuring a stronger mechanical structure and improved calibration, enhancing action precision and system robustness. We conduct comprehensive analyses to evaluate the system's efficiency and accuracy in demonstration collection.

Beyond scaling up in-the-wild demonstration collection, developing a robust policy is key to fully leveraging these demonstrations. We introduce \textbf{\textit{RISE}-2}, a generalizable policy that fuses 2D and 3D perception, as illustrated in Fig.~\ref{fig:teaser} (right). \textbf{\textit{RISE}-2} outperforms prior imitation policies in both in-domain and generalization settings. Notably, it matches the performance of teleoperated-data-trained policies using only in-the-wild demonstrations collected and adapted by \textbf{\textit{AirExo}-2} --- without any robot data. Under equivalent data collection time, it even surpasses teleoperated baselines. These results highlight the strong potential of combining \textbf{\textit{AirExo}-2} and \textbf{\textit{RISE}-2} for scalable, generalizable imitation learning. The hardware and codebase are open-sourced at \href{https://airexo.tech/airexo2/}{the project website}.

\section{Related Works}\label{sec:related-works}

\paragraph{Scaling up Demonstration Collection.} Demonstration data is fundamental to advancing generalizable imitation learning in robotic manipulation. Recent studies on data scaling laws~\cite{data_scaling_law} reveal that imitation learning in robotics follows similar scaling trends observed in natural language processing~\cite{scaling_nlp} and computer vision~\cite{scaling_video, dit} fields, highlighting the importance of scaling up demonstration collection. Teleoperation is the most direct and widely used approach for collecting real-world robot demonstrations, as it captures data in the robot's native observation domain and action space. However, scaling up teleoperated demonstration collection is challenging, primarily limited by the high cost and infrastructure demands of deploying more robots. Consequently, constructing large-scale teleoperated robotic datasets~\cite{rt1, agibotworld, oxe, rh20t, bcz, droid, bridgedata, robomind} requires substantial human labor and physical resources. Moreover, teleoperation is often less intuitive than natural human manipulation, resulting in higher user learning costs~\cite{hajl} and lower demonstration quality~\cite{s2i}. 

To overcome these limitations, recent research explores generating~\cite{gensim2, robotwin, gensim, robogen} or augmenting~\cite{rocoda, intervengen, dexmimicgen, mimicgen, cyberdemo} demonstrations automatically in simulation, which facilitates demonstration collection but often necessitates sim-to-real adaptation for real-world deployment. 
Other approaches leverage internet-scale human videos for representation learning~\cite{vrb, liv, vip, vc1, r3m, mvp, hrp, mpi} or policy pre-training~\cite{gen2act, gr2, vpdd, dexmv, gr1, lapa}, yet accurately converting 2D human videos into robot demonstrations with 3D trajectories and robot states remains a challenge~\cite{human_video_survey}. Alternatively, in-the-wild data collection uses specialized human-operated devices --- such as hand-held grippers~\cite{umi, rum_stickv2, legato, dobbe_stickv1, demoat}, cameras~\cite{ar2d2, eve}, VR/AR systems~\cite{arcap, egomimic, m2r}, motion-capture gloves~\cite{dexcap}, and exoskeletons~\cite{airexo_v1, m2r} --- to translate human motions into robot actions. 

\paragraph{Learning from In-the-Wild Demonstrations.} Despite promising in scalability, the kinematic gap and the visual gap pose obstacles in learning from in-the-wild demonstrations~\cite{airexo_v1, m2r}. \textbf{(1)} The kinematic gap refers to discrepancies in motion translation between humans and robots, which can lead to inaccuracies in action execution. This gap can be addressed through visual or mechanical methods. Visual approaches typically use off-the-shelf pose estimation frameworks from commercial devices~\cite{arcap, umi, ar2d2, rum_stickv2, egomimic, legato, dobbe_stickv1, dexcap, eve} or structure-from-motion~\cite{sfm, demoat} to extract robot poses from demonstrations. In contrast, mechanical approaches~\cite{airexo_v1, m2r} employ kinematically isomorphic devices that capture robot poses through angle encoder readings. \textbf{(2)} The visual gap, on the other hand, arises from the presence of specialized devices and human hands in the captured visual data, whereas robot demonstrations must accurately reflect the robot itself. Hand-held devices~\cite{umi, rum_stickv2, legato, dobbe_stickv1, demoat} mitigate this by using in-hand cameras and the same end-effector during deployment to maintain visual consistency. Other methods address the visual gap by incorporating robot demonstrations for fine-tuning or co-training~\cite{ar2d2, airexo_v1, egomimic, eve}, or by employing human-in-the-loop strategies to collect corrective behaviors during policy deployment~\cite{arcap, dexcap}.

\paragraph{Generalizable Manipulation Policy.} Direct learning from in-the-wild demonstrations highlights the need for a generalizable manipulation policy capable of transferring learned skills to the robot during real-world deployment. A generalizable policy can adapt to new domains or environments, even with limited training data from restricted domains~\cite{cage}, and should generalize across variations in camera perspectives, backgrounds, objects, and embodiments. While many behavior-cloning-based robotic manipulation policies~\cite{bc, dp, act3d, rvt, peract, act} perform well in-domain, they often fail in out-of-distribution scenarios, leading to compounded errors and task failures~\cite{rise, cage}. Large-scale pre-training on real-world robot demonstration data can improve generalization~\cite{mt-act, rt1, oxe, bcz, openvla, rdt1b, octo, hpt, rt2}, but it does not fully address the inherent limit of the policy in adapting across diverse domains. Two promising approaches for improving generalization involve leveraging 3D perceptions~\cite{polarnet, act3d, rvt, rise, dp3} to enhance environmental understanding and using 2D foundation models~\cite{what_pretrain, spawnnet, soft, cage, same} to enrich the policy's recognition of complex object and scene information. Recent work~\cite{lift3d, sgr, sgrv2} combines these approaches to further enhance policy performance, though challenges remain in effectively integrating these strategies for broader generalization across diverse manipulation domains.

\section{Method}

\subsection{\textit{AirExo}-2: Efficiently Collecting and Adapting In-the-Wild Demonstrations}\label{sec:airexo2}

As discussed in \S\ref{sec:related-works}, eliminating the need for physical robots enables low-cost and scalable in-the-wild demonstration collection. However, visual and kinematic discrepancies introduce significant challenges for directly utilizing such data. To overcome this, our objective is to \textit{efficiently convert in-the-wild demonstrations into pseudo-robot demonstrations suitable for training real-world manipulation policies}. This requires \textbf{(1)} robust hardware for reliable data collection, \textbf{(2)} accurate motion tracking, and \textbf{(3)} photo-realistic visual adaptation to support effective policy learning.

\begin{figure*}
    \centering
    \includegraphics[width=\linewidth]{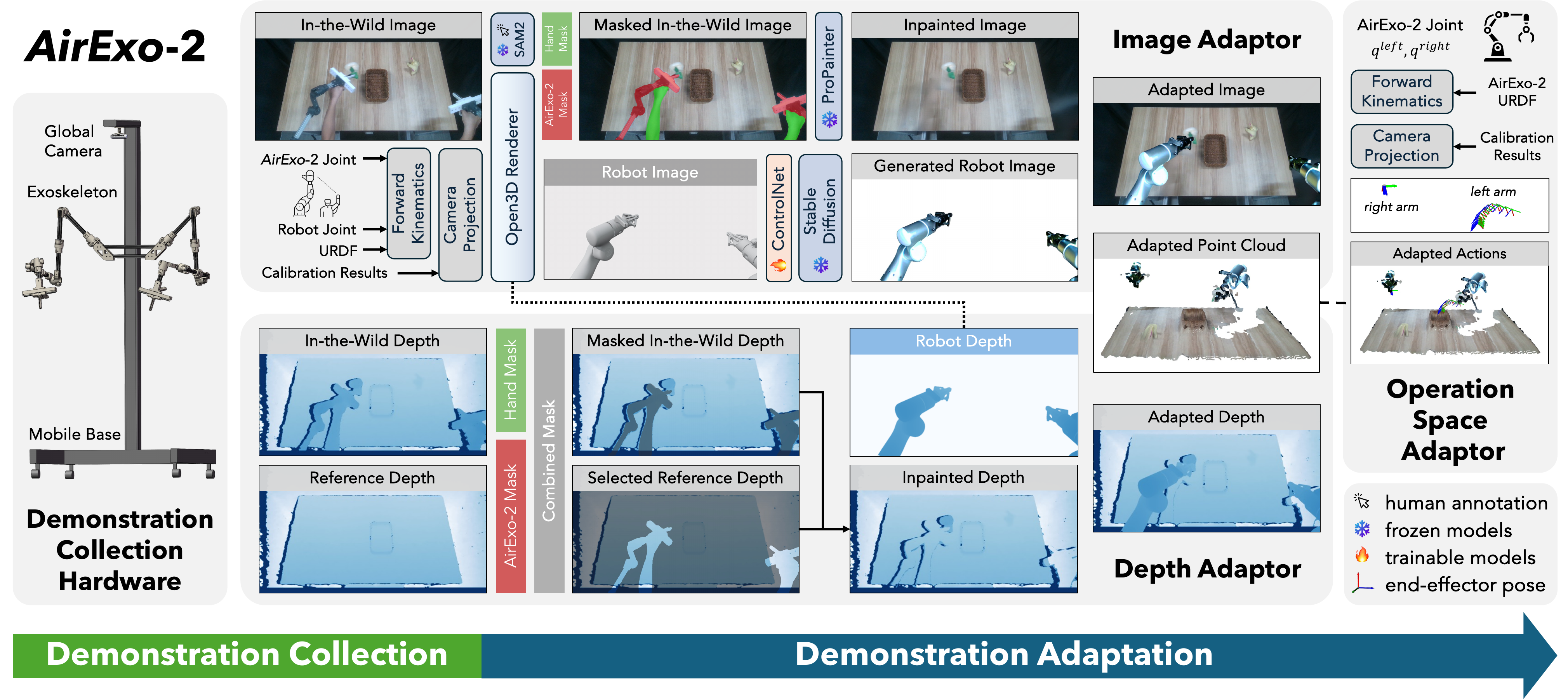}
    \caption{\textbf{Overview of the \textit{AirExo-2} System}. (left) Human uses the low-cost mobile \textbf{demonstration collection device} for data collection. (top) The \textbf{image adaptor} transforms the RGB images into the robot domain. (bottom) The \textbf{depth adaptor} transforms the depth images into the robot domain. (right) The \textbf{operation space adaptor} generates the states and actions based on the hardware encoder readings.}
    \label{fig:airexo2}
\end{figure*}

To this end, we present the \textbf{\textit{AirExo}-2} system, designed to efficiently collect and adapt in-the-wild demonstrations for downstream manipulation policy training, as illustrated in Fig.~\ref{fig:airexo2}. Unlike prior approaches that \textit{circumvent} the visual gap through in-hand cameras~\cite{umi, rum_stickv2, legato, dobbe_stickv1, demoat} or rely on pre-training and fine-tuning paradigm~\cite{arcap, ar2d2, airexo_v1, egomimic, dexcap, eve}, our system explicitly \textit{bridges} this gap by converting in-the-wild observations into pseudo-robot views using visual adaptors. Combined with accurate motion tracking enabled by the mechanical design, this semi-automatic adaptation process better aligns demonstrations with robot deployment conditions, allowing policies to \textit{directly} transfer to real-world robot platform. Please refer to Appendix~\ref{app:airexo2} for implementation details.

\paragraph{Hardware.} As shown in Fig.~\ref{fig:airexo2} (left), we upgrade the AirExo design~\cite{airexo_v1} with aluminum profiles to improve structural robustness and motion tracking precision. To facilitate subsequent demonstration adaptations, we design the \textbf{\textit{AirExo}-2} to have a 1:1 scale and maintain kinematic isomorphism with the robotic arm. A mobile base is added to support data collection in diverse environments, and a dedicated calibration process is introduced to ensure accurate motion capture. Details of the hardware design and calibration procedure are provided in Appendix~\ref{app:airexo2-hardware} and Appendix~\ref{app:airexo2-calibration}, respectively. The system retains the low-cost advantage of the original AirExo, with the dual-arm platform (excluding the camera) priced at just \$600.

\paragraph{Operation Space Adaptor.} Following~\cite{rise, idp3}, we project all states and actions into the global camera coordinate --- a universal coordinate frame between the \textbf{\textit{AirExo}-2} and robot platform.

\paragraph{Image Adaptor.} As shown in Fig.~\ref{fig:airexo2} (top), we convert in-the-wild image into pseudo-robot image by (1) inpainting human-related regions --- including hands and \textbf{\textit{AirExo}-2} --- and (2) overlaying it with the generated robot image. Leveraging the joint encoder readings, calibration results, and the joint mapping between \textbf{\textit{AirExo}-2} and the robot, we render RGB-D and mask images of both systems in Open3D~\cite{open3d}. The rendered \textbf{\textit{AirExo}-2} mask is merged with the hand mask obtained via SAM-2~\cite{sam2} to localize human-related areas, which are then inpainted using ProPainter~\cite{propainter}. Inspired by~\cite{rovi_aug}, the rendered robot image is refined via ControlNet-guided Stable Diffusion~\cite{stable_diffusion, controlnet} and composited onto the inpainted image to generate the final photo-realistic pseudo-robot image.

\paragraph{Depth Adaptor.}
As shown in Fig.~\ref{fig:airexo2} (bottom), we first capture a universal reference depth of the empty workspace, and combine it with the static object depth from the first frame to form a reference depth for each demonstration. Guided by the combined mask from the image adaptor, we inpaint human-related regions in the depth map with the reference depth to maintain spatial consistency. The adapted depth is obtained by merging the inpainted depth with the rendered robot depth.

\subsection{\textit{RISE}-2: A Generalizable Policy for Learning from In-the-Wild Demonstrations}\label{sec:rise2}

\begin{figure*}
    \centering
    \includegraphics[width=\linewidth]{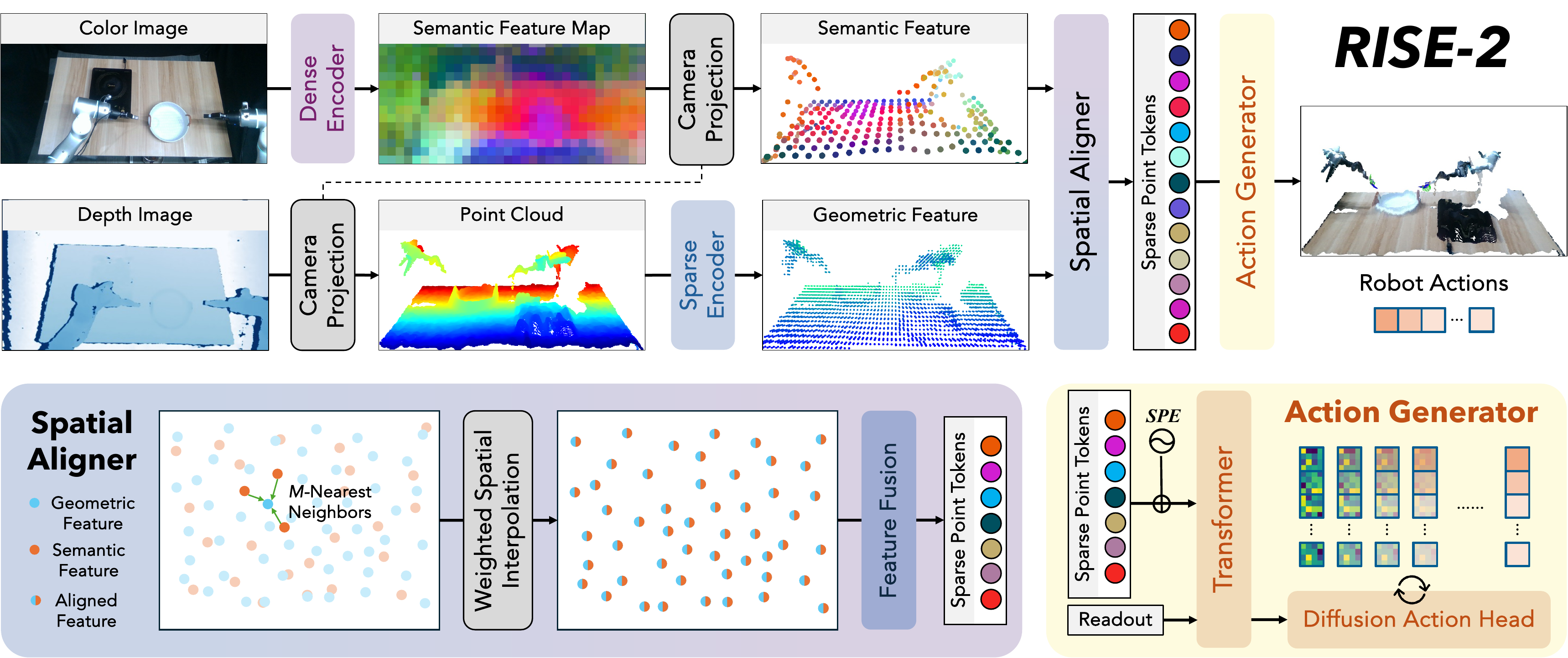}
    \caption{\textbf{\textit{RISE}-2 Policy Architecture}. Given an RGB-D observation as input, the policy predicts continuous actions in the camera frame. The architecture consists of four key components: (1) the \textbf{dense encoder} to obtain 2D semantic features; (2) the \textbf{sparse encoder} to obtain the 3D local geometric features of seed points; (3) the \textbf{spatial aligner} for the feature fusion of 2D semantic features and 3D geometric features guided by their 3D coordinates; (4) the \textbf{action generator} to generate continuous robot actions based on the fused feature.}
    \label{fig:policy}
\end{figure*}

Although in-the-wild demonstrations are adapted into pseudo-robot demonstrations, domain gaps --- such as differences in camera viewpoints --- still persist. To effectively learn from such data, a generalizable policy is required. As discussed in \S\ref{sec:related-works}, 3D perception provides view-invariant geometric information, while 2D foundation models offer rich semantic understanding from large-scale pretraining. To leverage the complementary strengths of both, we introduce \textbf{\textit{RISE}-2}, a generalizable policy that integrates 3D spatial perception with 2D semantic features for robust robotic manipulation. As illustrated in Fig.~\ref{fig:policy}, the \textbf{\textit{RISE}-2} policy consists of a \textit{sparse encoder} for 3D features, a \textit{dense encoder} for 2D features, a \textit{spatial aligner} for feature fusion, and an \textit{action generator} that maps fused features to actions. Implementation details are provided in Appendix~\ref{app:policy}.

\paragraph{Sparse Encoder.} 3D point cloud data contains rich spatial structure, enabling effective extraction of local geometric features --- a property widely used in general grasping~\cite{anygrasp,fang2020graspnet,wang2021graspness}. RISE~\cite{rise} encodes point clouds using raw color information to capture both semantic and geometric cues, but it struggles to disentangle coordinate shifts from color variations. \textbf{\textit{RISE}-2} inherits the sparse 3D encoder from RISE, but removes color input to focus purely on geometric features. Formally, the sparse encoder $\mathrm{E}_s: (D, K) \rightarrow (\mathbf{F}_g, \mathbf{C}_g)$, maps a depth image $D$ and camera intrinsics $K$ to sparse geometric features $\mathbf{F}_g$, along with the corresponding seed point coordinates $\mathbf{C}_g = \{c_g^i \in \mathbf{P}\}$ after network down-sampling. 
The encoder adopts a ResNet-like architecture~\cite{resnet} built on MinkowskiEngine~\cite{minkowski}, enabling efficient sparse feature extraction and supporting real-time policy execution. 

\paragraph{Dense Encoder.} A generalizable policy requires rich semantic features to understand the scene, while the low-quality texture of point cloud data poses a challenge to this demand. Hence, \textbf{\textit{RISE}-2} adopts a dense 2D encoder to focus on semantic understanding based on the color image. Formally, the dense encoder $\mathrm{E}_d: (I, D, K) \rightarrow (\mathbf{F}_s, \mathbf{C}_s)$ processes the color image $I$ into semantic feature map $\mathbf{F}_s=\{f_s^i\}$ densely organized in the 2D form of shape $h\times w$, along with its reference 3D coordinates $\mathbf{C}_s = \{c_s^i\}$, calculated by applying 2D adaptive average pooling to size $h \times w$ on the point cloud generated from the depth map $D$ and camera intrinsics $K$. One key advantage of the dense encoder is its compatibility with 2D foundation models, which provides highly generalizable visual representations across diverse tasks and domains~\cite{cage}. We implement $\mathrm{E}_d$ using DINOv2~\cite{dino} fine-tuned with LoRA~\cite{lora}, enhancing the robustness and adaptability of the policy in capturing contextual relationships within the environment.

\paragraph{Spatial Aligner.} \textbf{\textit{RISE}-2} extracts geometric and semantic features using separate encoders, posing a challenge for feature fusion from different domains. A straightforward solution is to concatenate the aggregated feature vectors, but this discards fine-grained local information critical for precise perception. An alternative~\cite{sgr,sgrv2} upsamples $\mathbf{F}_s$ to match the resolution of the input image $I$, projects it onto the point cloud, and then downsamples to align with the seed points $\mathbf{C}_g$. However, this process is computationally expensive and reduces the efficiency of policy training and deployment. Instead, \textbf{\textit{RISE}-2} utilizes a spatial aligner to efficiently fuse two kinds of features based on their 3D coordinates $\mathbf{C}_g$ and $\mathbf{C}_s$. For each point $c_g^i\in\mathbf{C}_g$ output by the sparse encoder $\mathrm{E}_s$, we compute its $M$ nearest neighbors $\mathbf{N}_i=\{n^i_1, n^i_2, \cdots, n^i_M\} \subseteq \mathbf{C}_s$ from the coordinates output by the dense encoder $\mathrm{E}_d$. The aligned semantic feature $f^i_{s*}$ of $c_g^i$ is computed by weighted spatial interpolation:
\begin{equation}
    f^i_{s*} = \frac{\Sigma_{j=1}^M f_s^j / \mathrm{dist}(c_g^i,n^i_j)}{\Sigma_{j=1}^M 1/\mathrm{dist}(c_g^i,n^i_j)},
\end{equation}
where $\mathrm{dist}(c_i, c_j)$ denotes the Euclidean distance between $c_i$ and $c_j$. Then, the aligned feature for point $c_s^i$ is obtained by concatenation, \textit{i.e.}, $f_i = \mathrm{Concat}(f_g^i, f_{s*}^i)$. By aligning the seed points $\mathbf{C}_g$ with the dense semantic feature map $\mathbf{F}_s$ through their 3D coordinates, we can accurately retrieve fine-grained semantic features at arbitrary locations. The fused features are then aggregated into high-level sparse representations using sparse convolution layers~\cite{minkowski}. Visualization of the weighted spatial interpolation applied to the 2D feature map is provided in Appendix~\ref{app:dino-feat}.

\paragraph{Action Generator.} \textbf{\textit{RISE}-2} utilizes a decoder-only transformer~\cite{vaswani2017attention} to map point tokens with sparse positional encoding~\cite{rise} to the latent feature, which conditions a diffusion head~\cite{dp,ddpm,janner2022planning} to generate the action chunk~\cite{act}. Actions are represented in the camera frame for cross-scene consistency, with translations in absolute coordinates and rotations in the 6D format~\cite{rot6d}.
\section{Experiments}\label{experiments}

\subsection{Policy Evaluation: \textit{RISE}-2}

\begin{figure*}
    \centering
    \includegraphics[width=\linewidth]{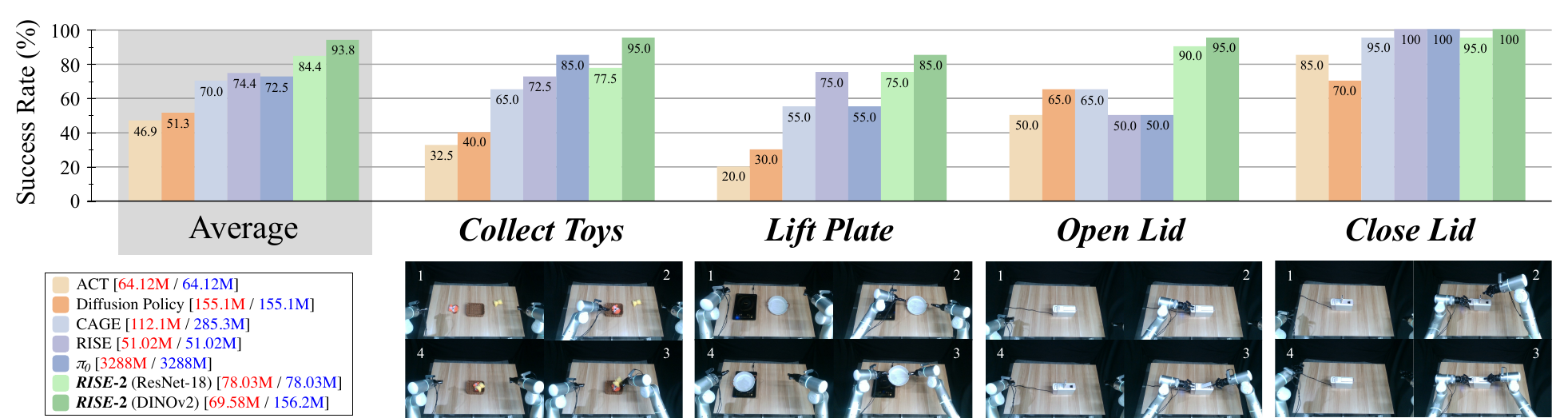}
    \caption{\textbf{Tasks and In-Domain Evaluation Results}. We design 4 tasks to evaluate the performance of the \textbf{\textit{RISE}-2} policy. Each policy is noted as ``policy [{\color{red}\# trainable parameters} / {\color{blue}\# total parameters}]''. \textbf{\textit{RISE}-2} outperforms previous state-of-the-art imitation policies while using relatively few parameters.}
    \label{fig:task}
\end{figure*}

\paragraph{Setup.} To evaluate the performance of policies in both in-domain and generalization settings, we design 4 tasks in Fig.~\ref{fig:task} and collect 50 demonstrations for each task using teleoperation~\cite{airexo_v1}. We compare \textbf{\textit{RISE}-2} against several representative policies based on 2D images or 3D point clouds, including ACT~\cite{act}, Diffusion Policy~\cite{dp}, CAGE~\cite{cage}, RISE~\cite{rise}, and $\pi_0$~\cite{pi0}. Additionally, we include a \textbf{\textit{RISE}-2} variant with the dense encoder implemented with an ImageNet-pretrained~\cite{imagenet} ResNet-18~\cite{resnet}, denoted as \textbf{\textit{RISE}-2} (ResNet-18). Please refer to Appendix~\ref{app:exp-setup} for details.

\paragraph{In-Domain Evaluation.} As shown in Fig.~\ref{fig:task}, \textbf{\textit{RISE}-2} significantly outperforms previous state-of-the-art imitation policies in average success rate across 4 tasks, demonstrating its effectiveness in handling diverse manipulation challenges. On tasks requiring precise motion execution, such as \textbf{\textit{Lift Plate}} and \textbf{\textit{Open Lid}}, \textbf{\textit{RISE}-2} achieves superior accuracy in predicting fine-grained actions, consistently surpassing all baselines. The performance gain over RISE~\cite{rise} highlights the benefits of using separate 2D and 3D encoders with spatial feature alignment. Further improvements from \textbf{\textit{RISE}-2} (ResNet-18) to \textbf{\textit{RISE}-2} (DINOv2) underscore the value of leveraging diverse knowledge from visual foundation models~\cite{dino} in robotic manipulation.

\begin{wraptable}{r}{0.4\linewidth}
\vspace{-0.6cm}
    \centering\fontsize{7}{8}\selectfont
    \setlength\tabcolsep{1.5pt}
    \begin{tabular}{ccrrr}
        \toprule
        \multirow{2}{*}{\textbf{Method}} & \multirow{2}{*}{\textbf{\textit{In-Domain}}} & \multicolumn{3}{c}{\textbf{\textit{Generalization}}} \\
        \cmidrule(lr){3-5}
        & & \multicolumn{1}{c}{Bg.} & \multicolumn{1}{c}{Obj.} & \multicolumn{1}{c}{Both} \\
        \midrule
        ACT~\cite{act} & 32.5\% & 0.0\% & 2.5\% & 0.0\% \\
        Diffusion Policy~\cite{dp} & 40.0\% & 12.5\% & 5.0\% & 2.5\% \\
        CAGE~\cite{cage} & 65.0\% & 45.0\% & 42.5\% & 10.0\% \\
        RISE~\cite{rise} & 72.5\% & 42.5\% & 12.5\% & 40.0\% \\ 
        $\pi_0$~\cite{pi0} & 85.0\% & 77.5\% & 82.5\% & \textbf{82.5}\% \\ \midrule
        \textbf{\textit{RISE}-2} (ResNet-18) & 77.5\% & 47.5\% & 47.5\% & 37.5\% \\
        \textbf{\textit{RISE}-2} \textit{(DINOv2)} & \textbf{95.0}\% & \textbf{85.0}\% & \textbf{85.0}\% & 62.5\% \\
        \bottomrule 
    \end{tabular}
    \caption{\textbf{Generalization Evaluation Results}. ``Bg.'' denotes novel backgrounds and ``Obj.'' denotes novel objects. Policy success rates are reported.}
    \label{tab:eval-teleop-general} \vspace{-0.6cm}
\end{wraptable}
\paragraph{Generalization Evaluation.} We select the \textbf{\textit{Collect Toys}} task to evaluate the generalization ability of different policies under varying environmental disturbances. Please refer to Appendix~\ref{app:exp-setup} for disturbance details. As shown in Tab.~\ref{tab:eval-teleop-general}, \textbf{\textit{RISE}-2} demonstrates strong robustness, maintaining high performance even under combined disturbances. While the \textbf{\textit{RISE}-2} (ResNet-18) variant performs slightly below the \textbf{\textit{RISE}-2} (DINOv2) model, it still generalizes well and outperforms most baselines. These results further validate our architectural choice of using separate encoders for 3D geometric and 2D semantic features, which effectively enhances policy robustness.

\subsection{System Evaluation: \textbf{\textit{AirExo}-2}}

\begin{wrapfigure}{r}{0.38\textwidth}
\vspace{-0.5cm}
    \centering
    \includegraphics[width=\linewidth]{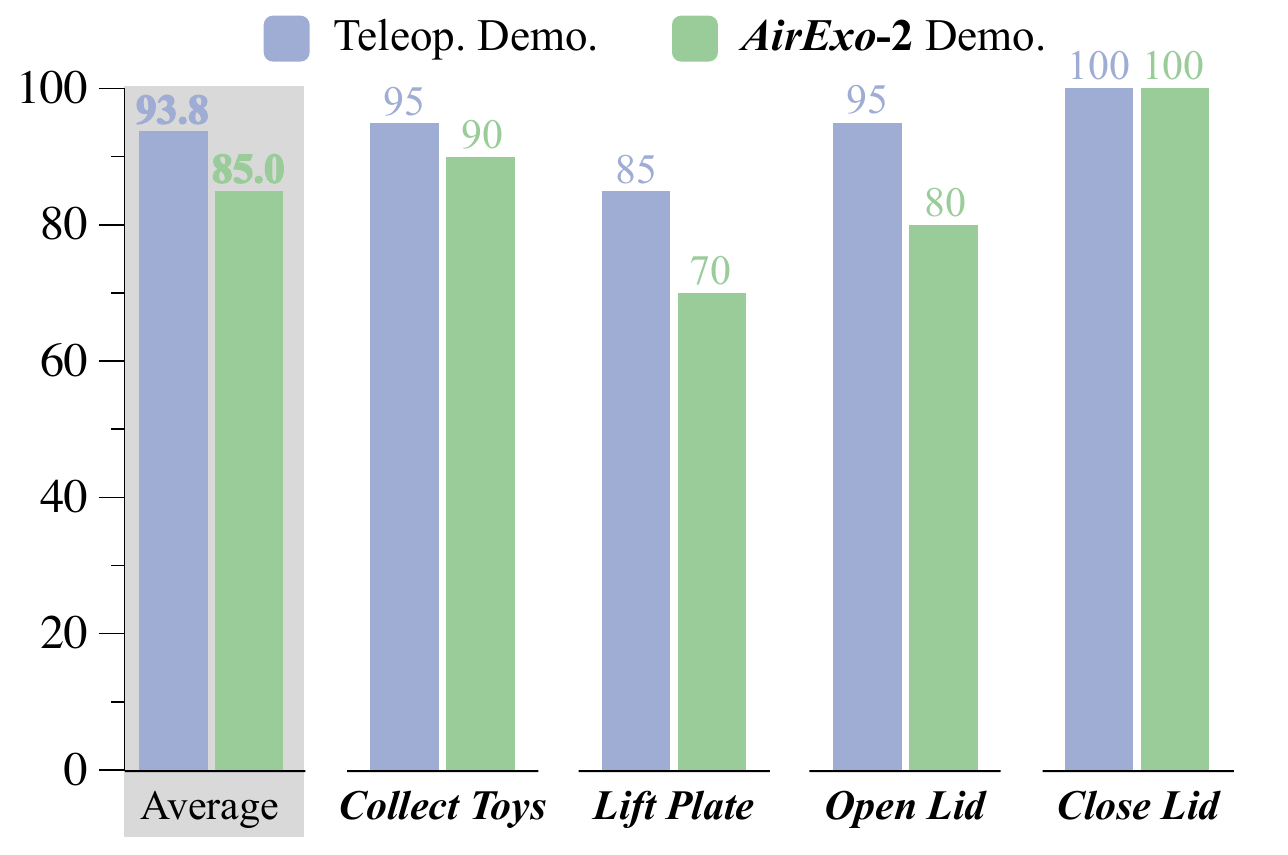}
    \caption{\textbf{\textit{RISE}-2 Policy Performance with Different Demonstrations}. Policies trained solely on demonstrations collected and adapted by \textbf{\textit{AirExo}-2} achieve reasonable performance without any robot data, demonstrating the overall effectiveness of the \textbf{\textit{AirExo}-2} system.}
    \label{fig:system}
    \vspace{-0.5cm}
\end{wrapfigure}
\paragraph{Learning from Pseudo-Robot Demonstrations.} Previous experiments have demonstrated that \textbf{\textit{RISE}-2} is a generalizable policy, making it well-suited for learning from in-the-wild demonstrations collected and processed by \textbf{\textit{AirExo}-2}. In this evaluation, we utilize \textbf{\textit{AirExo}-2} to collect an equal number of in-the-wild demonstrations and convert them into pseudo-robot demonstrations. These demonstrations are then used to train the \textbf{\textit{RISE}-2} policy, which is deployed on a real robot \textit{without further fine-tuning}. As shown in Fig.~\ref{fig:system}, the policy achieves satisfactory success rates on all tasks, with only a slight performance drop compared to the policy trained on teleoperated demonstrations. This highlights the ability of the \textbf{\textit{AirExo}-2} system to collect and adapt high-quality demonstrations suitable for training downstream imitation policies. Please refer to Appendix~\ref{app:additional-exp} for additional results of the RISE policy~\cite{rise}.

\begin{wrapfigure}{r}{0.3\textwidth}
    \vspace{-0.6cm}
    \centering
    \includegraphics[width=0.95\linewidth]{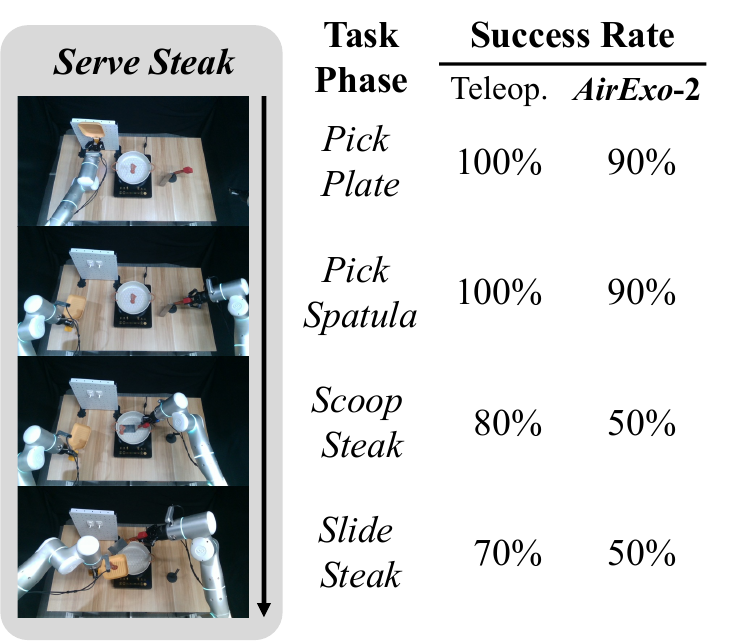}
    \caption{\textbf{Results of the \textit{Serve Steak} Task}. Trained only with demonstrations by \textbf{\textit{AirExo}-2}, the \textbf{\textit{RISE}-2} policy achieves decent performance in the challenging task, even without robot data.}
    \label{fig:complex}    
    \vspace{-0.4cm}
\end{wrapfigure}
\paragraph{Complex Task.} We evaluate the \textbf{\textit{AirExo}-2} system and the \textbf{\textit{RISE}-2} policy on the challenging, long-horizon, and contact-rich task \textbf{\textit{Serve Steak}}. Using 50 in-the-wild demonstrations collected and adapted by \textbf{\textit{AirExo}-2}, the \textbf{\textit{RISE}-2} policy achieves a decent 50\% overall success rate, with strong performance even in the contact-rich \textit{Scoop Steak} phase, as shown in Fig.~\ref{fig:complex}. This demonstrates the effectiveness of \textbf{\textit{AirExo}-2} in capturing accurate, fine-grained actions and the ability of \textbf{\textit{RISE}-2} to accomplish complex tasks without requiring robot data. Combined with the scalability analysis in \S\ref{sec:analysis}, increasing the number of in-the-wild demonstrations is expected to further boost performance. These results highlight the strong synergy between scalable, high-quality demonstration collection by \textbf{\textit{AirExo}-2} and generalizable \textbf{\textit{RISE}-2} policy learning, paving the way for efficient and adaptable robotic systems in real-world environments.

\begin{wraptable}{r}{0.36\textwidth}
    \vspace{-0.4cm}\footnotesize
    \centering
    \setlength\tabcolsep{2pt}
    \begin{tabular}{ccc}
        \toprule
        \multirow{2}{*}{\textbf{Method}} & \multicolumn{2}{c}{\textbf{Success Rate} $\uparrow$}  \\\cmidrule(lr){2-3}
        & \textit{w.o.} adaptors & \textit{w.} adaptors  \\
        \midrule
        RISE~\cite{rise}  & 30.0\% & 57.5\% \\
        \textbf{\textit{RISE}-2} & 52.5\% & 90.0\% \\
        \bottomrule 
    \end{tabular}
    \vspace{-0.1cm}
    \caption{\textbf{Ablation Results on Visual Adaptors}. The visual adaptors help bridge the embodiment visual gap, enabling effective policy transfer from in-the-wild demonstrations.}
    \label{tab:adaptor}
    \vspace{-0.4cm}
\end{wraptable}
\paragraph{Ablation on Visual Adaptors.} We use the \textbf{\textit{Collect Toys}} task to highlight the importance of the demonstration adaptor for learning from in-the-wild data. As shown in Tab.~\ref{tab:adaptor}, removing visual adaptors significantly reduces performance, revealing a substantial visual gap between in-the-wild and robot demonstrations. These results also validate the quality of \textbf{\textit{AirExo}-2} pseudo-robot demonstrations for training high-performing policies. Notably, \textbf{\textit{RISE}-2} without adaptors still performs comparably to RISE~\cite{rise} with adaptors, demonstrating its strong generalization across embodiments.

\subsection{System Analysis: \textit{AirExo}-2}\label{sec:analysis}

\begin{wraptable}{r}{0.5\linewidth}
    \vspace{-0.4cm}
    \centering\footnotesize
    \setlength\tabcolsep{1.5pt}
    \begin{tabular}{crccc}
    \toprule
        \multirow{2}{*}{\textbf{Method}} & \multirow{2}{*}{\begin{tabular}{c}\textbf{Completion}\\\textbf{Time} (s) $\downarrow$\end{tabular}} & \multicolumn{2}{c}{\textbf{Avg. Rank} $\downarrow$} & \multirow{2}{*}{\textbf{Rating} $\uparrow$}\\ \cmidrule(lr){3-4}
        & & \textbf{Intui.} & \textbf{Learn.} & \\
         \midrule
         EE Teleop. & $46.06_{\pm 27.21}$ & 3.00/3 & 2.95/3 & 29.75\\ 
         Joint Teleop. & $17.31_{\pm 5.055}$ & 1.80/3 & 2.00/3 & 49.58\\
         \textbf{\textit{AirExo}-2} & $\mathbf{5.66}_{\pm 1.978}$ & \textbf{1.20}/3 & \textbf{1.05}/3 & \textbf{83.00}\\
         \bottomrule
    \end{tabular}\vspace{-0.15cm}
    \caption{\textbf{User Study Results}. ``Intui.'' and ``Learn.'' denote intuitiveness and learnability, respectively.}
    \label{tab:study}
    \vspace{-0.3cm}
\end{wraptable}
\paragraph{User Study.} We compare three demonstration collection methods: end-effector pose teleoperation using a haptic device~\cite{rh20t}, joint-space teleoperation via AirExo~\cite{airexo_v1}, and in-the-wild collection with \textbf{\textit{AirExo}-2}. Twenty participants (14 men, 6 women, aged 21–35) of varying expertise levels perform the \textbf{\textit{Collect Toys}} task on all three platforms \textit{in randomized order}, with each trial consisting of a 3-minute familiarization and a timed demonstration collection. After completing all trials, participants fill out a questionnaire to provide feedback on three collection methods. From the results in Tab.~\ref{tab:study}, we observe that both experienced and novice users find \textbf{\textit{AirExo}-2} easier to learn and operate than both types of teleoperation, leading to faster onboarding and smoother operation. Statistical analysis (Welch's $t$-test, $p = 8.32\times 10^{-10}$) further confirms its superior efficiency in task completion time. The user-friendliness and efficiency make \textbf{\textit{AirExo}-2} well-suited for diverse users and consistent, high-quality data collection at scale in real-world settings.

\begin{wraptable}{r}{0.37\textwidth}
\vspace{-0.4cm}
    \centering\footnotesize
    \setlength\tabcolsep{4pt}
    \begin{tabular}{crr}
        \toprule
        \multirow{2}{*}{\textbf{Device}} & \multicolumn{2}{c}{\textbf{Error} (mm) $\downarrow$} \\ \cmidrule(lr){2-3}
        & \multicolumn{1}{c}{Avg $_{\pm\ \text{Std}}$} & \multicolumn{1}{c}{Max} \\
        \midrule
        UMI~\cite{umi} & $8.855_{\pm 3.228}$ & 20.002 \\
        \textbf{\textit{AirExo}-2} & $\textbf{1.737}_{\pm 1.713}$ & \ \textbf{6.134} \\
        \bottomrule 
    \end{tabular}\vspace{-0.1cm}
    \caption{\textbf{Action Accuracy of Data Collection Systems}. \textbf{\textit{AirExo}-2} achieves higher action accuracy than handheld devices like UMI~\cite{umi}.}
    \label{tab:itw-accuracy}
    \vspace{-0.8cm}
\end{wraptable}
\paragraph{Action Accuracy.}
Accurate actions are essential for learning reliable policies from in-the-wild demonstrations. Unlike handheld devices such as UMI~\cite{umi}, which use visual SLAM for trajectory estimation, \textbf{\textit{AirExo}-2} leverages its mechanical structure and forward kinematics for more precise motion tracking. As shown in Tab.~\ref{tab:itw-accuracy}, it achieves significantly lower translation errors across 3 tracks (please refer to Appendix~\ref{app:accuracy-analysis} for details about accuracy evaluations), confirming its reliability for high-fidelity motion capture.

\begin{wrapfigure}{r}{0.38\textwidth}
\vspace{-0.6cm}
\centering
    \includegraphics[width=\linewidth]{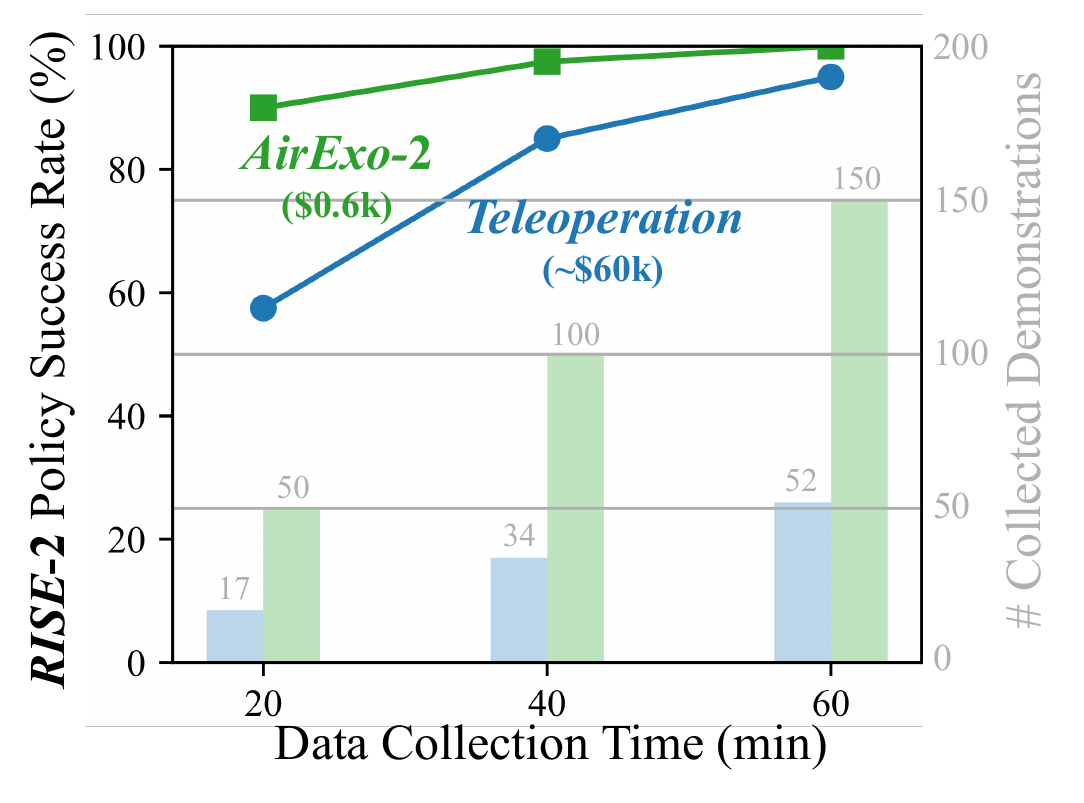}
    \vspace{-0.5cm}
    \caption{\textbf{Scalability Analysis Results}. \textbf{\textit{AirExo}-2} enables scalable and efficient demonstration collection without compromising data effectiveness.}
\label{fig:scalability}
\vspace{-0.5cm}
\end{wrapfigure}

\paragraph{Scalability Analysis.} We compare the scalability and effectiveness of \textbf{\textit{AirExo}-2} in-the-wild data collection with traditional teleoperation data collection in Fig.~\ref{fig:scalability}. Given the same collection time, \textbf{\textit{AirExo}-2} yields higher demonstration throughput (analogous to the findings in the previous user study) with lower costs (\$0.6k per \textbf{\textit{AirExo}-2} system versus approximately \$60k per teleoperation platform with robots), making it a more scalable solution to demonstration collection. Moreover, \textit{using the demonstrations collected under the same time}, policies trained on \textbf{\textit{AirExo}-2} demonstrations outperform those trained on teleoperation data, underscoring the value of adapted pseudo-robot demonstrations for efficient and effective imitation learning.

\section{Conclusion}\label{sec:conclusion}

This paper introduces \textbf{\textit{AirExo}-2}, a novel system designed for scalable in-the-wild demonstration collection and adaptation using low-cost exoskeletons. By incorporating several adaptors, \textbf{\textit{AirExo}-2} enables the visual and kinematic transformation of in-the-wild demonstrations into pseudo-robot demonstrations, which can then be directly applied to downstream imitation learning tasks. We also propose a generalizable policy, \textbf{\textit{RISE}-2}, which effectively integrates both 2D and 3D perception, demonstrating exceptional performance in both in-domain and generalization scenarios.

Further experiments on a variety of tasks demonstrate that when trained exclusively on pseudo-robot demonstrations generated by the \textbf{\textit{AirExo}-2} system --- without using any robot data --- the policy achieves satisfactory performance during zero-shot deployment on a real-world robot platform, even for the complex, long-horizon, and contact-rich task \textbf{\textit{Serve Steak}}. This highlights the potential of combining \textbf{\textit{AirExo}-2} and \textbf{\textit{RISE}-2} as a scalable and promising alternative to traditional teleoperation-imitation pipelines, providing a more efficient, cost-effective solution for scalable, generalizable robotic imitation learning. Together, these results open new possibilities for transferring manipulation skills from in-the-wild environments to real robots, without the need for extensive robot-centric data collection.
\section{Limitations and Future Works}\label{sec:limitation}

While we utilize the proposed demonstration adaptor to visually transform in-the-wild demonstrations collected by \textbf{\textit{AirExo}-2} into pseudo-robot demonstrations, these transformed demonstrations are primarily useful for generalizable policies. To enhance the applicability of the pseudo-robot demonstrations to a broader range of policies, future work could explore the integration of demonstration augmentation methods, such as novel view synthesis~\cite{rovi_aug, zeronvs, vista, gcd}, into the adaptation process to improve the diversity of the demonstrations, making them more versatile for various policy learning.

As demonstrated by several works~\cite{umi, inhand_than_global, giving_robots_a_hand, demoat}, the in-hand image is a semi-unified observation modality across different embodiments. Our case study in Appendix~\ref{app:case-study} also reveals that combining in-hand observations can strengthen the performance of various 2D policies. However, the current \textbf{\textit{AirExo}-2} system does not include in-hand cameras. Although we have designed connectors to integrate them with the exoskeleton, calibrating the in-hand cameras with the exoskeleton remains challenging, making it difficult to adapt the in-hand images into the robot domain. Future work could explore effective methods for adapting in-hand images collected by \textbf{\textit{AirExo}-2} to the robot domain, or investigate strategies for leveraging the semi-unified in-hand observations in policy design.

The current \textbf{\textit{AirExo}-2} system only supports parallel grippers as end-effectors, limiting its applicability in more dexterous tasks. Future work could integrate the \textbf{\textit{AirExo}-2} system with dexterous hands~\cite{eyesight_hand, leap_hand} and their corresponding exoskeletons~\cite{dexos, ben2025homie}, enabling more complex manipulation capabilities and expanding the range of tasks the system can effectively perform. Moreover, adding additional sensors to the \textbf{\textit{AirExo}-2} system for recording other observation modalities --- such as force/torque~\cite{foar, forcemimic}, tactile~\cite{vitamin}, and sound~\cite{maniwav} --- can also be explored in the future.

\acknowledgments{We would like to thank Peishen Yan and Chunyu Xue from Shanghai Jiao Tong University for their valuable suggestions on the figures in this paper. We would like to thank Teng Hu from Shanghai Jiao Tong University for insightful discussions about inpainting strategies to mitigate the visual gap. 

This work was supported by the National Key Research and Development Project of China (No. 2022ZD0160102), the National Key Research and Development Project of China (No. 2021ZD0110704), Shanghai Artificial Intelligence Laboratory, XPLORER PRIZE grants.}

\printbibliography

\clearpage
\appendix
{\LARGE \bf Appendix}
\section{\textbf{\textit{AirExo}-2} Implementation}\label{app:airexo2}

\subsection{Hardware Design}\label{app:airexo2-hardware}

High-quality in-the-wild data requires precise collection for effective pseudo-robot transformation. However, AirExo~\cite{airexo_v1} has several key hardware limitations:

\begin{enumerate}
    \item[\textbf{H1}.] Most of its components are 3D-printed using polylactic acid‌ (PLA), leading to low rigidity and susceptibility to structural deformation.
    \item[\textbf{H2}.] Unlike typical robots with constrained ranges, its joints can rotate beyond 360$^\circ$, potentially reaching positions the robot cannot achieve.
    \item[\textbf{H3}.] Although portable, inevitable body movements during operations can cause its base to shift, leading to inaccurate action recording.
    \item[\textbf{H4}.] Its gripper lacks smooth control, leading to potential jamming under clamping forces.
    \item[\textbf{H5}.] The shaft connects directly to the encoder, with side-routed wires that endure friction and stretching during joint movement, risking long-term damage or short circuits.
\end{enumerate}

During demonstration collection using teleoperation, most of the above issues (except \textbf{H4}) might not significantly impact data acquisition, as the human operator can adjust their actions based on the movements of the robotic arms. However, during the in-the-wild demonstration collection, these drawbacks can substantially affect motion capture accuracy and may result in invalid or unusable demonstration data.

\begin{figure}[h]
    \centering
    \includegraphics[width=0.65\linewidth]{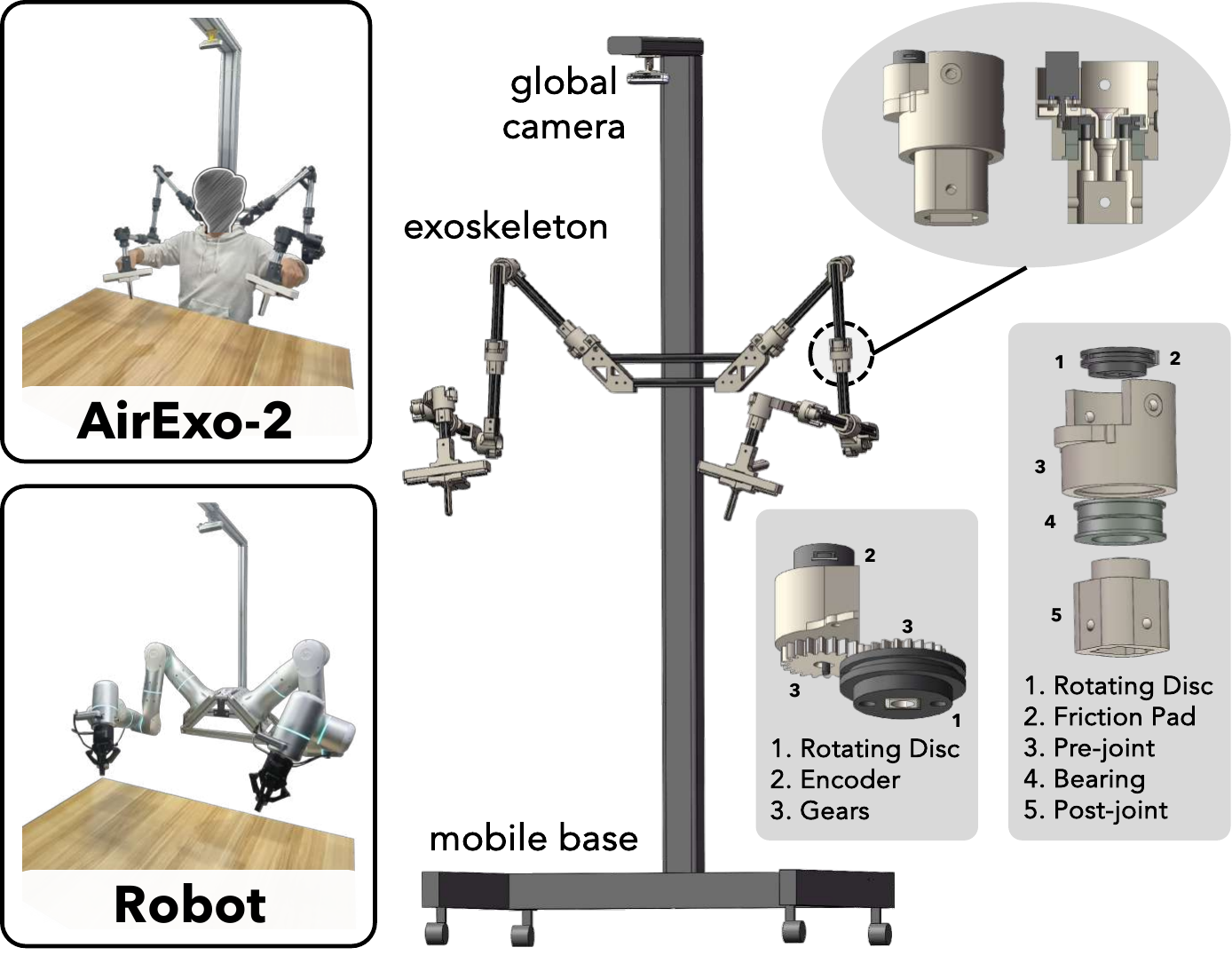}
    \caption{\textbf{Hardware Design of \textit{AirExo}-2}. The \textbf{\textit{AirExo}-2} demonstration collection platform consists of a mobile base and a dual-arm exoskeleton. Global cameras can be mounted on top of the platform to capture visual observations during data collection. The detailed joint structure is shown on the right side, featuring two key designs: hollow rotating discs and side-mounted encoders; joints with angle limit and adjustable friction.}
    \label{fig:hardware}
\end{figure}

Driven by the limitations above, the hardware design of \textbf{\textit{AirExo}-2} focuses on seamless integration with the learning process from in-the-wild demonstrations. To enable direct learning, the exoskeleton is built with a 1:1 dimensional match to the robotic arm, minimizing discrepancies between human and robot embodiments. An overview of the hardware is shown in Fig.~\ref{fig:airexo2}. Excluding cameras, the complete \textbf{\textit{AirExo}-2} platform costs \$600. Key hardware design elements are outlined below.

\paragraph{Enhanced Overall Structural Rigidity (H1).} In AirExo, the links connecting two consecutive joints are the most prone to deformation. In \textbf{\textit{AirExo}-2}, we replace the 3D-printed parts with 20x20 European standard aluminum profiles, providing significantly higher strength at a very low cost. For the joints, the outer shell is 3D-printed using PLA-CF, a carbon fiber-reinforced PLA material with higher hardness. Inside the joint, larger bearings are used to further enhance structural rigidity. Together with the improved links, the hardware upgrade significantly increases the overall structural rigidity of \textbf{\textit{AirExo}-2}, making the exoskeleton more durable, and thereby improving system accuracy.

\paragraph{Hollow Rotating Disc and Side-Mounted Encoder (H5).} As shown in Fig.~\ref{fig:hardware}, this design features an encoder mounted on the side of the joint, which uses a gear mechanism to translate the rotational angle of the rotating disc into encoder readings. The hollow disc design allows wires to pass through, preventing them from stretching during rotation and thereby extending their lifespan. Additionally, the side-mounted encoder simplifies maintenance, enabling easy debugging and replacement without the need to disassemble the joint.

\paragraph{Joint with Angle Limit and Adjustable Friction (H2).} This structure consists of a rotating disc with a grooved track and a friction pad that can be embedded into the track. Together, they allow for adjustment of the limiting angle, ensuring that the motion range of the \textbf{\textit{AirExo}-2} joint exactly aligns with the corresponding joint range of the robot. As illustrated in Fig.~\ref{fig:hardware}, the rotating disc, pre-joint, and post-joint are connected through bearings, allowing the rotational motion of the joint to be directly transmitted to the track. For demonstration collection, excessive or insufficient friction in the rotation of the joints is undesirable. Hence, the friction force of the joint in \textbf{\textit{AirExo}-2} can be adjusted by turning the screw on the outer shell, which compresses the friction pad. This design ensures optimal friction for comfortable and accurate data collection.

\paragraph{Smooth Gripper Control (H4).} Following~\cite{umi, forcemimic}, the gripper of \textbf{\textit{AirExo}-2} incorporates a linear guide, with the fingers mounted on a sliding block. This design allows for smoother opening and closing of the gripper, ensuring it operates seamlessly under clamping forces without any stalling.

\paragraph{Mobile Data Collection Platform (H3).} Portability is crucial for in-the-wild data collection. However, to address the issue of base movement caused by the body motion of the operator, we mount \textbf{\textit{AirExo}-2} on a mobile aluminum profile stand, as shown in Fig.~\ref{fig:hardware}. This setup ensures stability of the base during demonstration collection while maintaining the flexibility needed for mobility, enabling large-scale demonstration collection in real-world environments. An Intel RealSense D415 camera is set up on the top of the mobile platform to capture global observations. We also designed two optional camera mounts (though not used in this paper) for the future integration of in-hand cameras on the top of both grippers.

\subsection{Calibration}\label{app:airexo2-calibration}

The \textbf{\textit{AirExo}-2} system requires two types of calibration simultaneously: (1) aligning the zero positions of each joint with the corresponding robot joint, and (2) determining the transformation between the global camera and the \textbf{\textit{AirExo}-2} base. To address these challenges, we propose a two-stage calibration process.

\paragraph{Initial Calibration.} For initial calibration, the former calibration can be achieved by manually adjusting the joints to approximate the zero position using specialized 3D-printed tools and reading the encoder values, obtaining $\{\tilde{q}_\text{calib}^{\text{left}}, \tilde{q}_\text{calib}^{\text{right}}\}$. The latter calibration can be done by attaching an ArUco calibration marker board with a known position on the base $\mathbf{T}^\text{base}_\text{marker}$ and performing optical calibration using OpenCV~\cite{opencv}, obtaining $\mathbf{T}^\text{camera}_\text{marker}$. Thus, the transformation can calculated as 
\begin{equation}
\left[\tilde{\mathbf{t}}^\text{camera}_\text{base} \mid \tilde{\mathbf{r}}^\text{camera}_\text{base}\right] \stackrel{\text{def}}{=} \tilde{\mathbf{T}}^\text{camera}_\text{base} = \mathbf{T}^\text{camera}_\text{marker}\left(\tilde{\mathbf{T}}^\text{base}_\text{marker}\right)^{-1}
\end{equation}

However, this approach introduces errors due to human observation, calibration board misalignment, and optical inaccuracies. In a chained system like \textbf{\textit{AirExo}-2}, these errors may propagate and amplify across joints, leading to significant cumulative inaccuracies in the end-effector pose. Therefore, fine-grained calibration is essential to ensure precise and consistent alignment between \textbf{\textit{AirExo}-2} and the camera frames during demonstration collection. 

\paragraph{Calibration via Differentiable Rendering.} In the second stage, inspired by prior works~\cite{easyhec, easyhec++, sam-rl}, we use differentiable rendering~\cite{differentiable_rendering} to refine the initial calibration. Training samples are obtained from a single human \textit{play} trajectory with \textbf{\textit{AirExo}-2}. Using the joint states and calibration parameters, we render the system mask and depth via a differentiable rendering engine~\cite{redner}. Calibration parameters are optimized by minimizing discrepancies between the rendered and annotated system masks, as well as between the rendered and observed depths. Pseudo-ground-truth masks are manually annotated with SAM-2~\cite{sam2}. This iterative refinement compensates for errors that accumulate across joints, ultimately improving the overall accuracy.

Specifically, we define $p$, the calibration parameters to be optimized, as
\begin{equation}
p \stackrel{\text{def}}{=} \{\Delta \mathbf{t}^\text{camera}_\text{base}, \mathbf{r}^\text{camera}_\text{base}, \Delta q_\text{calib}^{\text{left}}, \Delta q_\text{calib}^{\text{right}}\}
\end{equation}
where parameters, except for the base-to-camera rotation, are represented as deltas relative to the initial calibration results. The base-to-camera rotation is expressed in a 6D format~\cite{rot6d}. Thus, the final calibration results can be calculated as
\begin{gather}\label{eq:recover-params1}
\mathbf{T}^\text{camera}_\text{base} = \left[\tilde{\mathbf{t}}^\text{camera}_\text{base} + \Delta \mathbf{t}^\text{camera}_\text{base} \mid \mathbf{r}^\text{camera}_\text{base} \right], \\
\label{eq:recover-params2}
q^{\text{type}}_\text{calib} = \tilde{q}^{\text{type}}_\text{calib} + \Delta q^{\text{type}}_\text{calib}, \quad \text{type}\in[\text{left}, \text{right}],
\end{gather}
and the initial parameter values are set as
\begin{equation}
p_0 = \{\mathbf{0}, \tilde{r}^\text{camera}_\text{base}, \mathbf{0}, \mathbf{0}\}
\end{equation}

\begin{figure}[t]
    \centering
    \includegraphics[width=0.9\linewidth]{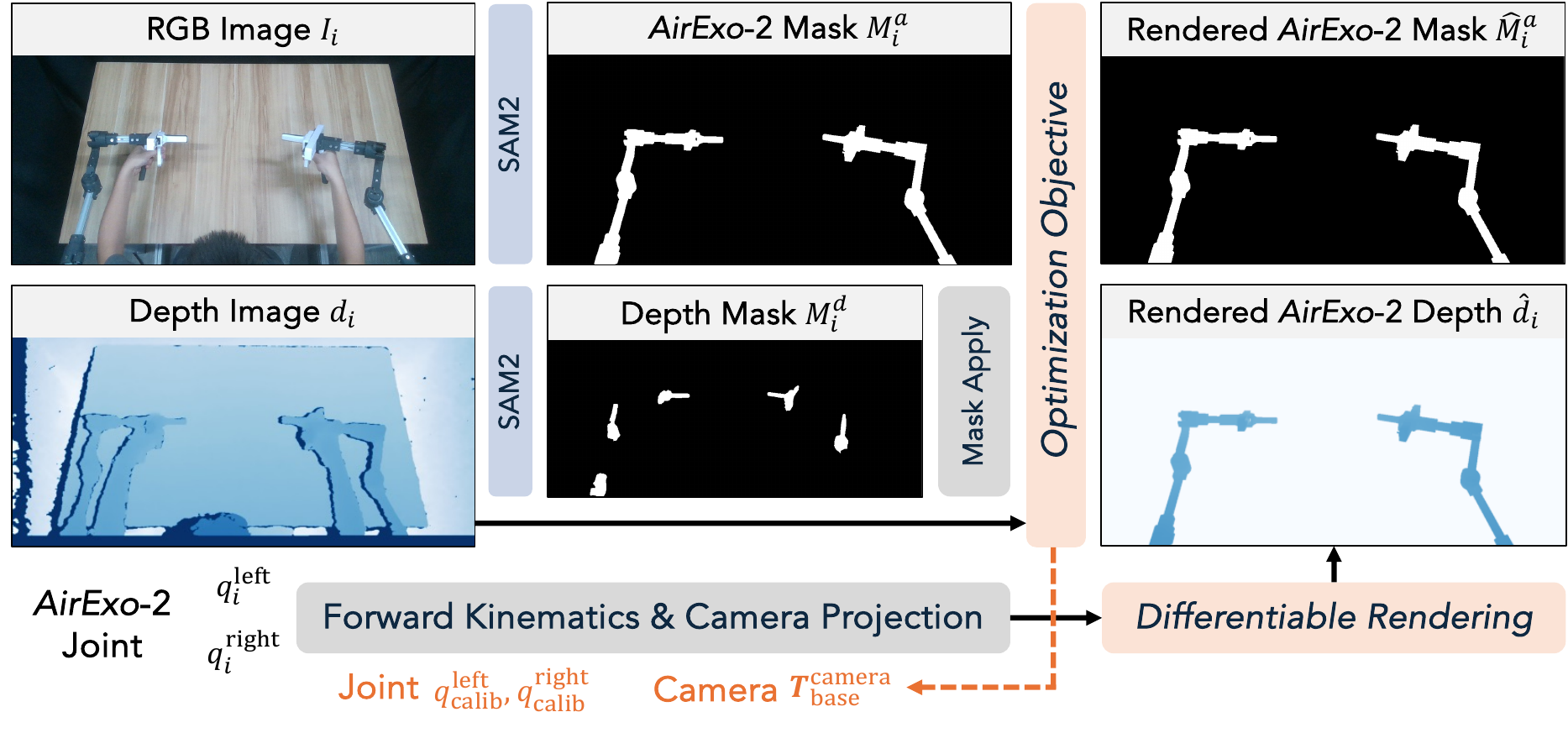}
    \caption{\textbf{Calibration via Differentiable Rendering}. The parameters in orange denote the calibration parameters to be optimized via differentiable rendering.}
    \label{fig:diff-calib}
    \vspace{-0.3cm}
\end{figure}

For the optimization process, we first record a single in-the-wild \textit{play} trajectory, in which the human operator uses the \textbf{\textit{AirExo}-2} to adopt various poses. During trajectory recording, ensure that all parts of the \textbf{\textit{AirExo}-2} remain above the human hands and arms from the camera's perspective. After data collection, we sample approximately 40 image-joint pairs from the trajectory, denoted as $\{I_i, d_i, q_i^\text{left}, q_i^\text{right}\}_{i=1}^{N_c}$, where $N_c$ represents the total number of training samples for calibration, and $I_i$ and $d_i$ are the RGB and depth images of the $i$-th sample, respectively. Subsequently, we utilize SAM-2~\cite{sam2} to annotate the \textbf{\textit{AirExo}-2} mask $M^a_i$ and the depth mask $M^d_i \subseteq M^a_i$ for each sample $i$, as shown in Fig.~\ref{fig:diff-calib}. The first mask provides supervision for the rendered \textbf{\textit{AirExo}-2} mask, while the second mask is used to select the valid depth information that serves as the supervision signal.

The differentiable rendering engine Redner~\cite{redner} is employed to render the \textbf{\textit{AirExo}-2} mask $\hat{M}_i^a$ and \textbf{\textit{AirExo}-2} depth $\hat{d}_i$ using the calibration results and joint information:
\begin{equation}
    \hat{M}_i^a, \hat{d}_i = \mathcal{R} (p; q_i^\text{left}, q_i^\text{right}, \tilde{T}_\text{base}^\text{camera},\tilde{q}^{\text{left}}, \tilde{q}^{\text{right}}),
\end{equation}
where the rendering engine $\mathcal{R}(p; \cdots)$ computes the gradients of the calibration parameters $p$ during the rendering process, and ${\tilde{T}_\text{base}^\text{camera}, \tilde{q}^{\text{left}}, \tilde{q}^{\text{right}}}$ represent the initial calibration results.

The rendered \textbf{\textit{AirExo}-2} mask $\hat{M}_i^a$ is supervised by the human-annotated pseudo-\textbf{\textit{AirExo}-2} mask $M^a_i$, and the rendered \textbf{\textit{AirExo}-2} depth $\hat{d}_i$ is supervised by the camera depth $d_i$ within the region of the human-annotated depth mask $M^d_i$. The depth mask ensures that only accurate depth information contributes to the loss. Thus, the objective can be written as:
\begin{equation}
    \mathcal{L}(p) = \frac{1}{N_c}\sum_{i=1}^{N_c}\left(\beta \cdot \left\|M^a_i - \hat{M}_i^a\right\|^2 + \left\| d_i - \hat{d}_i\right\|^2 \circ M^d_i \right),
\end{equation}
where $\beta$ represents the weighting coefficient, and $\circ$ denotes the mask-apply operation. In practice, we set $\beta = 5$, use $N_c = 40$ samples for optimization, and employ the Adam optimizer~\cite{adam} with a learning rate of $10^{-4}$ for 1000 iterations to fine-tune the calibration results.

\begin{wraptable}{r}{0.6\textwidth}
    \vspace{-0.4cm}
    \centering\footnotesize
    \setlength\tabcolsep{2pt}
    \begin{tabular}{ccc}
    \toprule
        \multirow{2}{*}{\textbf{Calibration Method}} & \multicolumn{2}{c}{\textbf{Difference} $\downarrow$} \\ \cmidrule(lr){2-3}
        & \textbf{Mask} (\%) & \textbf{Depth} (mm)\\
         \midrule
         Initial Calibration & $1.71_{\pm 0.37}$ & $21.6_{\pm 5.2}$\\
         Human Annotation & $2.31_{\pm 0.31}$ & $31.2_{\pm 6.4}$ \\
         Two-Stage Calibration (mask only) & $1.10_{\pm 0.26}$ & $17.6_{\pm 4.1}$ \\ 
         Two-Stage Calibration (mask + depth) & $\textbf{0.78}_{\pm 0.25}$ & $\textbf{14.0}_{\pm 2.9}$  \\
         \bottomrule
    \end{tabular}
    \vspace{-0.1cm}
    \caption{\textbf{Calibration Analysis Results}. Using our proposed two-stage calibration, we achieve higher accuracy than both initial calibration and human annotations. Including depth as additional supervision also helps the optimization process convergence.}
    \label{tab:calib}
    \vspace{-0.4cm}
\end{wraptable}

\paragraph{Calibration Analysis.} Precise calibration is vital for accurate action tracking. In addition to our proposed full two-stage calibration method, we evaluate three alternatives: (1) \textit{initial calibration}, using first-stage results without refinement; (2) \textit{human annotation}, where a user manually adjusts calibration parameters via a real-time GUI; and (3) \textit{two-stage calibration (mask only)}, which refines calibration using only mask differences as supervision. As shown in Tab.~\ref{tab:calib}, our proposed full two-stage method achieves the highest accuracy, with a 0.78\% mask difference and 14.0 mm depth error. While depth errors are affected by sensor noise and cannot fully reflect action accuracy (please refer to \S\ref{sec:analysis} for action accuracy analysis), the results confirm the benefits of modeling 3D information through differentiable rendering for accurate calibration. Notably, human annotation even performs worse than initial calibration due to its reliance on 2D visual cues, which limits depth estimation accuracy.

\subsection{Visual Adaptors}

\paragraph{Semi-Automatic SAM-2 Annotations.} In the image adaptor, we initially annotate the hand mask (and, if visible, the head mask) manually using SAM-2~\cite{sam2}. After a few annotations on approximately 50 scenes, we can fine-tune SAM-2 on human-annotated samples, enabling automated labeling. This significantly reduces human effort and streamlines the demonstration adaptation process, making it nearly fully automated.

\paragraph{ControlNet Training.} We train a ControlNet~\cite{controlnet} based on the Stable Diffusion 1.5~\cite{stable_diffusion} model to generate photo-realistic robot images from rendered robot images. To collect training samples, we use teleoperation to gather a small amount of \textit{play} data, where the robot is teleoperated to move randomly within an empty workspace while recording RGB-D images and corresponding joint states. This ensures a diverse dataset of robot arm configurations, free from occlusions or distractions.

Notably, these training samples are \textbf{\textit{platform-specific}} but \textbf{\textit{task-invariant}}, meaning they only need to be collected once per robot platform, and the trained ControlNet can be used across all tasks. This also opens up the possibility of directly transforming our in-the-wild demonstrations to other robotic arms without the need to design new exoskeletons that match specific robots.

For training, we use a batch size of 88 and a learning rate of $10^{-5}$, while keeping other hyperparameters at their default settings. We use 50 DDPM sampling steps~\cite{ddpm} with a guidance scale of 9.0~\cite{cfg}. The prompt for generating robot images for our robot platform is displayed as follows.

\begin{tcolorbox}[colback=gray!5, colframe=black, boxrule=0.5pt, arc=2mm,
    left=2mm, right=2mm, top=1mm, bottom=1mm, fontupper=\small\ttfamily, width=\linewidth]
\textbf{Prompt:} robotic arms, dual arm, industrial robotic manipulator, metallic silver color, mechanical joints, precise mechanical details, gripper end effector, high-quality photo, photorealistic, clear and sharp details
\end{tcolorbox}

\subsection{Discussions}

\paragraph{Hardware Improvements over AirExo~\cite{airexo_v1}.} We improved both the \textit{repairability} and \textit{accuracy} of the system to enable scalable, high-quality data collection. In terms of repairability, relocating the encoders outside the joint and adopting a gear-driven design reduced encoder replacement time from 15 minutes to just 2 minutes. In terms of accuracy, the system's action recording precision was significantly enhanced, as presented in Tab.~\ref{tab:itw-accuracy}.

\paragraph{Computation Overhead of Adaptation.} Rendering a 1-minute scene of approximately 600 frames requires about 40 seconds on a single GPU, with the majority of the time consumed by the ControlNet-tuned Stable Diffusion~\cite{stable_diffusion, controlnet} model. This process can be further accelerated by: (1) rendering at a lower resolution, and (2) leveraging multiple GPUs in parallel. Notably, \textit{the primary bottleneck in data collection lies in human effort rather than computation}. Given that computational resources can be scaled more readily, we argue that prioritizing the reduction of human involvement --- even at the cost of slightly increased computational usage --- can significantly improve the overall efficiency and scalability of the data collection pipeline.

\paragraph{Cross-Robot Transfer and Data Reusability.} Our adaptation pipeline can be easily extended to other robots by deriving joint values from end-effector poses via inverse kinematics, followed by rendering and inpainting the corresponding robot. As shown in Tab.~\ref{tab:adaptor}, even without visual adaptation, policies trained solely on in-the-wild demonstrations can sometimes complete tasks successfully. This suggests that raw in-the-wild data itself can already be used to train policies and deployed on different robots. By incorporating robot-specific adaptations, we can generate high-quality data for different robots and boost policy performance by training on these adapted demonstrations. This approach can greatly enhance data reusability across platforms and embodiments.

\paragraph{Strict 1:1 Joint Mapping Design of \textit{AirExo}-2.} In our experiments, the Flexiv arms are comparable in size to the commonly used robotic arms like Franka, KUKA, and xArms. As shown in Tab.~\ref{tab:study}, the user study indicates that operators find our system intuitive and easy to use, suggesting that exact 1:1 scaling is not a prerequisite for effective data collection. Moreover, as most humanoid robots are designed to match human dimensions, our system is naturally compatible. For manipulators of different sizes, human-sized exoskeletons can still be employed, with data transfer facilitated through the adaptation process described in the previous discussion.

\section{\textbf{\textit{RISE}-2} Implementation}\label{app:policy}

\subsection{Data Processing}

\paragraph{Image.} The color image is resized to 448$\times$252 for DINOv2 backbone~\cite{dino} and 640$\times$360 for ResNet-18 backbone~\cite{resnet}. The depth image is resized to 640$\times$360 before creating the point cloud. The camera intrinsics are adjusted accordingly. Both the point clouds and actions are projected to the camera coordinate system following~\cite{rise,idp3}.

\paragraph{Point Cloud.} The point cloud is down-sampled with a voxel size of 5mm. For the data collected with teleoperation, we crop the point clouds using the range of 
$$
(x,y,z) \in [-0.70\textrm{m}, 0.70\textrm{m}] \times [-0.30\textrm{m}, 0.55\textrm{m}] \times [0.90\textrm{m}, 1.55\textrm{m}].
$$

For the data collected with \textbf{\textit{AirExo}-2}, we crop the point clouds using the range of 
$$
(x,y,z) \in [-0.70\textrm{m}, 0.70\textrm{m}] \times [-0.30\textrm{m}, 0.45\textrm{m}] \times [0.75\textrm{m}, 1.4\textrm{m}].
$$

\paragraph{Robot Trajectory Sampling.} The robot trajectories are sampled using differences of translation, rotation, and gripper width to remove redundant actions. For the action at two adjacent timesteps, if all the differences are less than the thresholds, only the first action is retained. The threshold for translation and gripper width is 5mm, and the rotation threshold is $\pi/24$.

\subsection{Hyperparameters}

\paragraph{Policy.} The sparse encoder adopts a ResNet-like architecture built upon MinkowskiEngine~\cite{minkowski}. The dense encoder adopts DINOv2-base~\cite{dino} as the 2D backbone with the output channel of 128. In the spatial aligner, we use $M=3$ for feature alignment. The aligned features are fused by shared MLPs with the size of (256, 256, 256), and then fed into another sparse network. The two sparse networks are detailed in Tab.~\ref{tab:rise-2-detail}. The transformer in action generator contains 4 blocks, in which we set $d_{\mathrm{model}} = 512$ and $d_{\mathrm{ff}} = 2048$ following~\cite{rise}. The channel number of the readout token is 512. The diffusion head adopts a CNN implementation~\cite{dp} with 100 denoising iterations in training and 20 iterations in inference. The output action horizon used in experiments is 20.

\begin{table}[h]
    \centering\footnotesize
    \begin{tabular}{ccc}
        \toprule
        \textbf{Layer Name} & \textbf{Sparse Encoder} & \textbf{Spatial Aligner} \\
        \midrule
         \multirow{3}{*}{init\_conv} & $k=[3,3,3]$, $c=32$, & \multirow{3}{*}{-} \\
         & $d=1$, $s=1$\\
         & 2x mean pooling \\
         \midrule
         \multirow{2}{*}{conv1} & $k=[3,3,3]$, $c=32$, & $k=[3,3,3]$, $c=256$,\\
         & $d=1$, $s=1$ & $d=4$, $s=4$ \\
         \midrule
         \multirow{2}{*}{conv2} & $k=[3,3,3]$, $c=64$, & $k=[3,3,3]$, $c=256$,\\
         & $d=2$, $s=1$ & $d=1$, $s=2$ \\
         \midrule
         \multirow{2}{*}{conv3} & $k=[3,3,3]$, $c=128$, & $k=[3,3,3]$, $c=512$,\\
         & $d=4$, $s=1$ & $d=1$, $s=2$ \\
         \midrule
         \multirow{2}{*}{conv4} & $k=[3,3,3]$, $c=128$, & $k=[3,3,3]$, $c=512$,\\
         & $d=8$, $s=2$ & $d=1$, $s=2$ \\
         \midrule
         \multirow{2}{*}{final\_conv} & $k=[1,1,1]$, $c=128$, & $k=[1,1,1]$, $c=512$,\\
         & $d=1$, $s=1$ & $d=1$, $s=1$ \\
        \bottomrule 
    \end{tabular}\vspace{0.2cm}
        \caption{\textbf{Sparse Convolutional Network Parameters of the \textit{RISE}-2 Policy}. Both the sparse encoder and the spatial aligner utilize MinkResNet~\cite{minkowski} for point cloud encoding. $k, c, d, s$ stand for the kernel size, output channel number, dilation, and stride in the convolutional layers, respectively.}
    \label{tab:rise-2-detail}
\end{table}

\paragraph{Training.} We train the \textbf{\textit{RISE}-2} policy on 4 NVIDIA A100 GPUs for 60k steps with a batch size of 240, an initial learning rate of $3\times 10^{-4}$, and a warmup step of $2000$. We employ a cosine scheduler to adjust the learning rate during training. During training, 20\% of the color images are augmented using color jitter with (brightness, contrast, saturation, hue) parameters set to (0.4, 0.4, 0.2, 0.1).

\paragraph{Inference.} We deploy \textbf{\textit{RISE}-2} using an RTX 2060 SUPER GPU and achieve a 5Hz inference speed, demonstrating efficient performance compared to larger vision-language-action policies~\cite{pi0}. We additionally provide scripts that enable policy inference to be executed on a remote server.

\subsection{Visualization of Sparse Semantic Features}\label{app:dino-feat}

\begin{figure}[h]
    \centering
    \includegraphics[width=\linewidth]{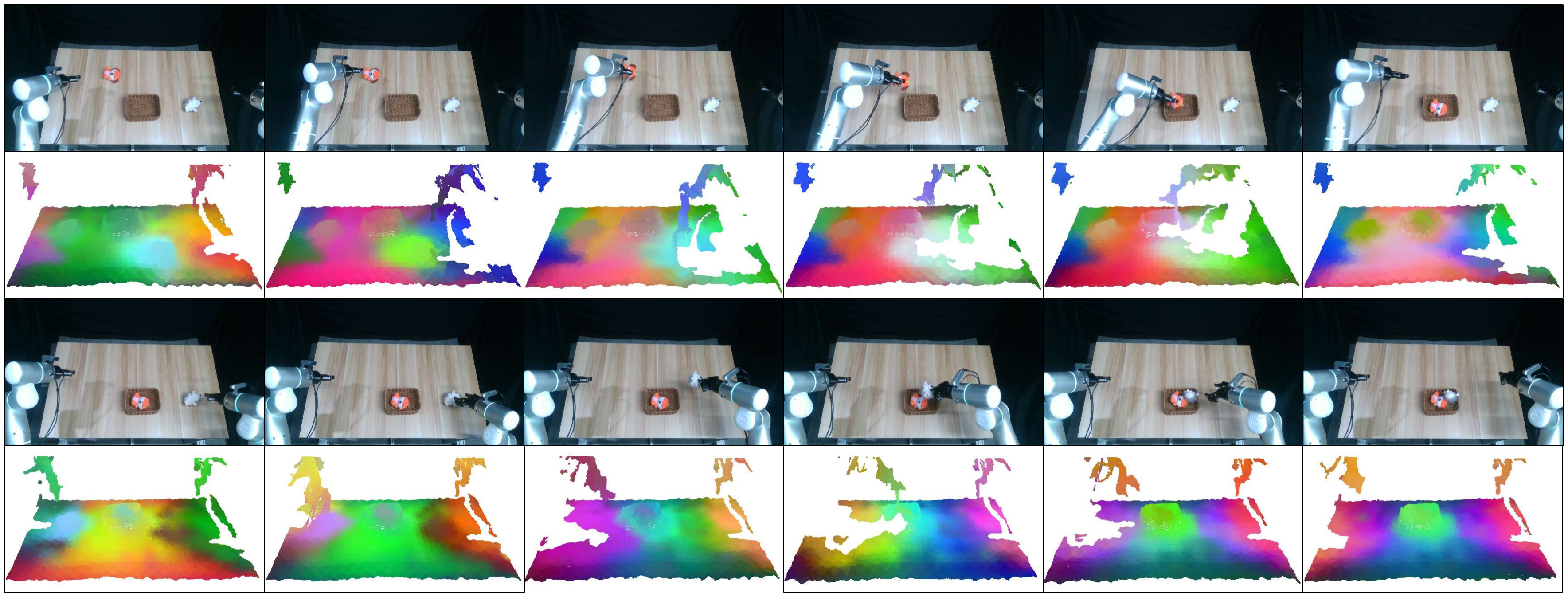}
    \caption{\textbf{Visualization of Sparse Semantic Features}. The colors are obtained by performing PCA on the features. The original sparse semantic features are aligned to the input point cloud using weighted spatial interpolation function in the spatial aligner for clearer visualization.}
    \label{fig:dino_feat}
\end{figure}

Fig.~\ref{fig:dino_feat} visualizes the sparse semantic features obtained from the dense encoder by projecting the 2D feature map to 3D form using the reference coordinates. The sparse semantic features are aligned to the input point cloud using weighted spatial interpolation, as detailed in \S\ref{sec:rise2}. Although the 2D feature map output by the dense encoder is in low resolution (32$\times$18 for DINOv2~\cite{dino} backbone), we still observe clear and distinguishable continuous feature variations on the aligned features, where the targets at the current step can be easily identified from the entire scene. Such a characteristic ensures precise feature fusion in the spatial domain. Additionally, we find that the features from the visual foundation model DINOv2~\cite{dino} change significantly as the task progresses, enabling the model to clearly understand the global state at the current time.

\section{Experiments Details}

\subsection{Setup}\label{app:exp-setup}

\paragraph{Data Collection.} The teleoperated demonstrations are collected using AirExo~\cite{airexo_v1}, while in-the-wild demonstrations are collected and adapted using our proposed \textbf{\textit{AirExo}-2} system. For each task, unless further specified, we collect 50 teleoperated demonstrations for \textbf{\textit{RISE}-2} policy evaluation and 50 in-the-wild demonstrations for \textbf{\textit{AirExo}-2} system evaluations.

\paragraph{Baselines.}  We compare \textbf{\textit{RISE}-2} against several representative policies based on 2D images and 3D point clouds, including:
(1) \textit{ACT}~\cite{act}, which employs transformers to map image observations and proprioception to robot action chunks;
(2) \textit{Diffusion Policy}~\cite{dp}, which formulates action prediction as a diffusion denoising process~\cite{ddpm, ddim} conditioned on the image observations;
(3) \textit{CAGE}~\cite{cage}, an extension of Diffusion Policy that incorporates visual foundation models~\cite{dino}, a causal observation perceiver~\cite{perceiver}, and an attention-based diffusion action head for improved generalization;
(4) \textit{RISE}~\cite{rise}, a 3D imitation policy that leverages a sparse 3D encoder for efficient point cloud perception;
and (5) $\pi_0$~\cite{pi0}, a vision-language-action flow model for general robot control.

\paragraph{Evaluation Protocols.} All policies except $\pi_0$ are deployed on a workstation with an NVIDIA RTX 2060 SUPER GPU, while $\pi_0$ is deployed remotely on a server with an NVIDIA A100 GPU for GPU memory considerations. Following the procedure outlined in~\cite{umi, cage}, we adopt a consistent evaluation method for each policy to minimize performance variation and ensure reproducibility. Specifically, we generate uniformly distributed test positions randomly before each task evaluation. The workspace is set up identically across different policies and test environments, and success rates are recorded for each test case. Each policy is evaluated over 20 consecutive trials per task (except for the \textbf{\textit{Serve Steak}} task, where we evaluate each policy for 10 consecutive trials), and the success rates are computed accordingly. 

\begin{wrapfigure}{r}{0.35\textwidth}
\vspace{-0.1cm}
    \centering
    \includegraphics[width=0.7\linewidth]{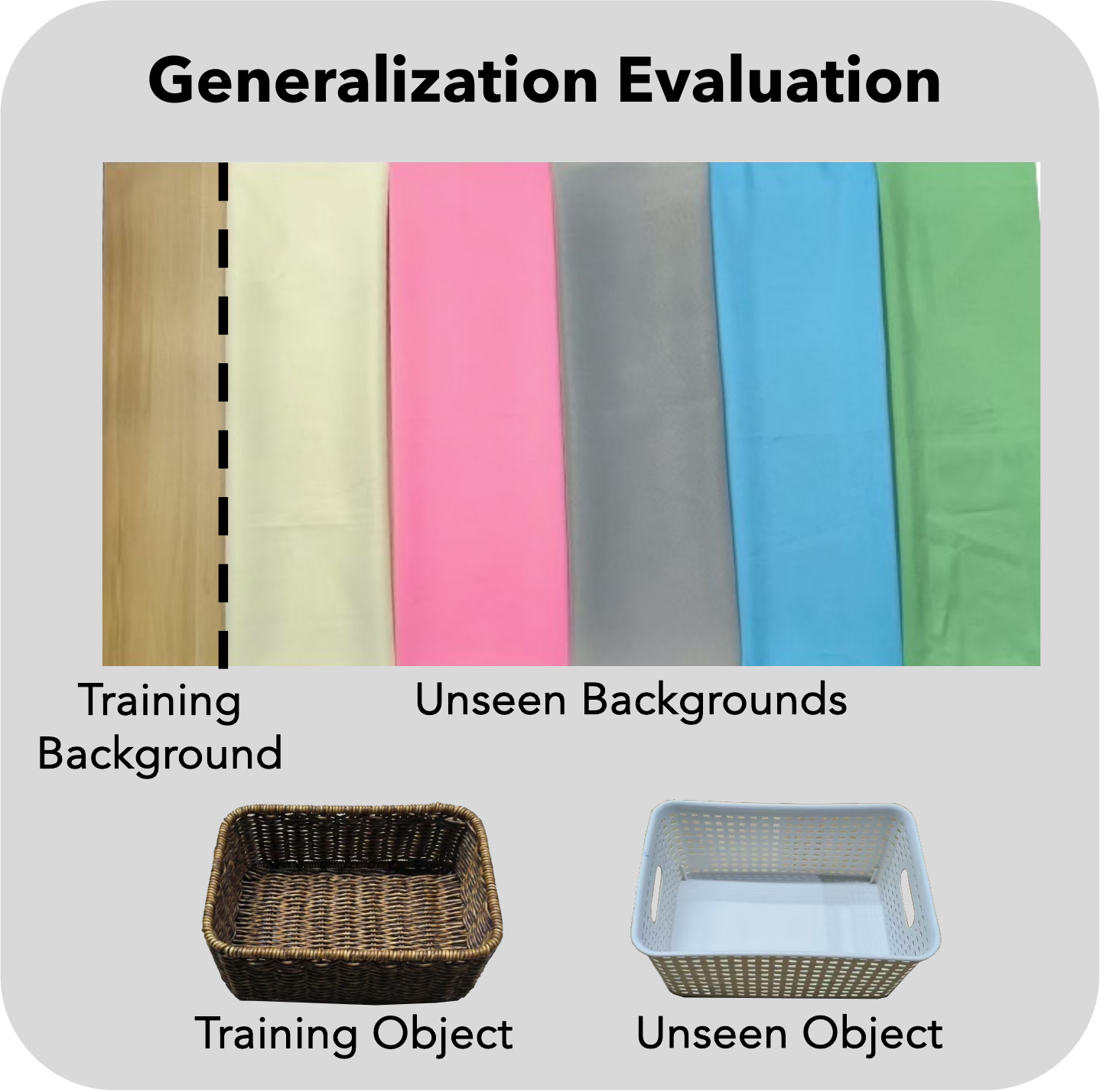}
    \caption{\textbf{Generalization Evaluation Setup}. We evaluate policy generalization using 5 unseen backgrounds and 1 unseen target object. Training is done in a narrow domain with only one background and one target object.}
    \label{fig:generalization}\vspace{-0.6cm}
\end{wrapfigure}
\paragraph{Generalization Evaluation.} We select the \textbf{\textit{Collect Toys}} task to conduct a generalization experiment to evaluate the robustness of different policies under varying levels of environmental disturbances. The policies are trained with demonstrations from a narrow domain, with only one training background and one target object. Then, we introduce two types of disturbances during generalization evaluations: background variations and object differences, as illustrated in Fig.~\ref{fig:generalization}. We then follow the same evaluation protocol introduced above to compute the policy success rates under (1) novel backgrounds, (2) novel objects, and (3) both novel backgrounds and novel objects, enabling a comprehensive comparison of each policy's generalization capability. For evaluation concerning the background variations, we use each of the 5 novel backgrounds for 4 evaluation trials in the same order to ensure fairness.

\subsection{Action Accuracy Analysis Details}\label{app:accuracy-analysis}

\begin{figure}[h]
\vspace{-0.2cm}
\centering
    \includegraphics[width=0.2\linewidth]{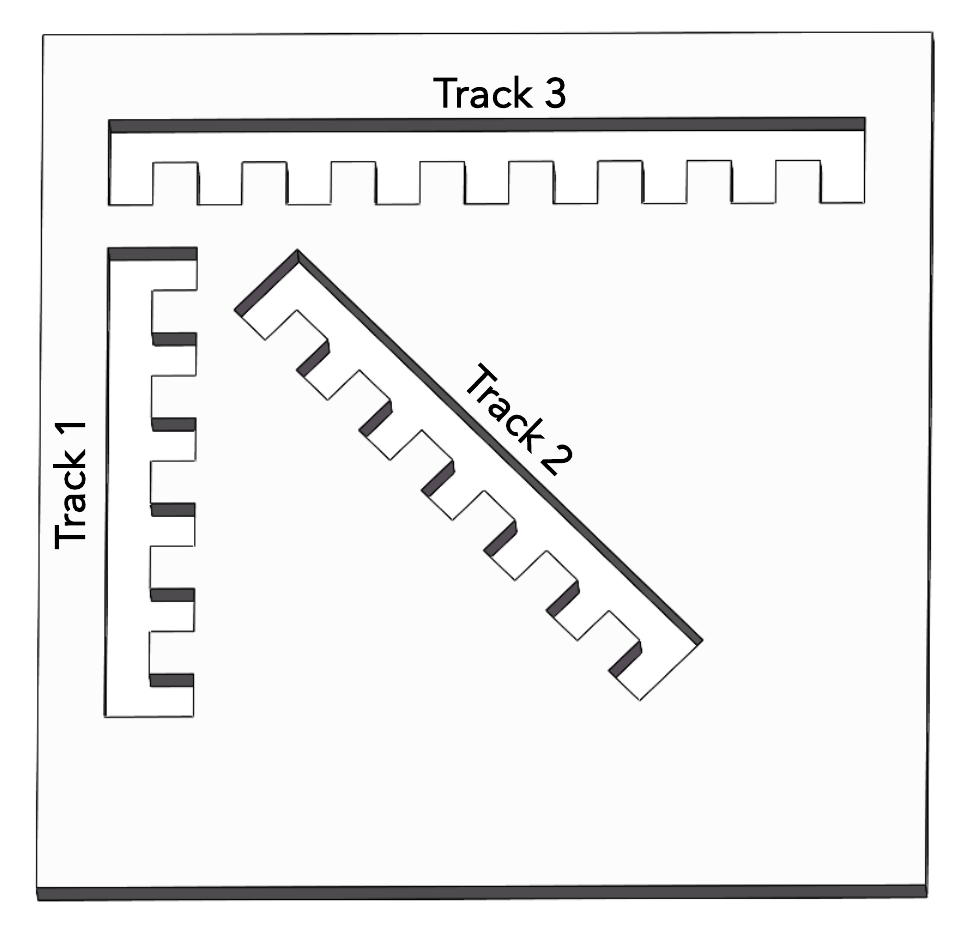}
    \vspace{-0.15cm}
    \caption{\textbf{The Evaluation Board}.}
\label{fig:track}\vspace{-0.4cm}
\end{figure}

To evaluate the action accuracy of the data collection systems, we designed a special evaluation board with three tracks, as illustrated in Fig.~\ref{fig:track}. Each track has fixed holes spaced 2 cm apart. We created custom connectors for both the \textbf{\textit{AirExo}-2} and UMI~\cite{umi} that fit into these fixed holes, allowing us to collect position data. By sequentially placing the connectors into each fixed hole along the track, we can calculate the relative movement distance between two adjacent fixed holes and compare it with the true value (20 mm) to calculate the error.

\subsection{Case Study: Are In-Hand Cameras Sufficient for Manipulations?}\label{app:case-study}

We conducted a case study to investigate whether in-hand cameras are sufficient for many manipulation tasks. We select the \textbf{\textit{Collect Toys}} task as an example and utilize CAGE~\cite{cage} as the policy for this case study. 

\begin{wrapfigure}{r}{0.4\textwidth}
\vspace{-0.8cm}
    \centering
    \includegraphics[width=\linewidth]{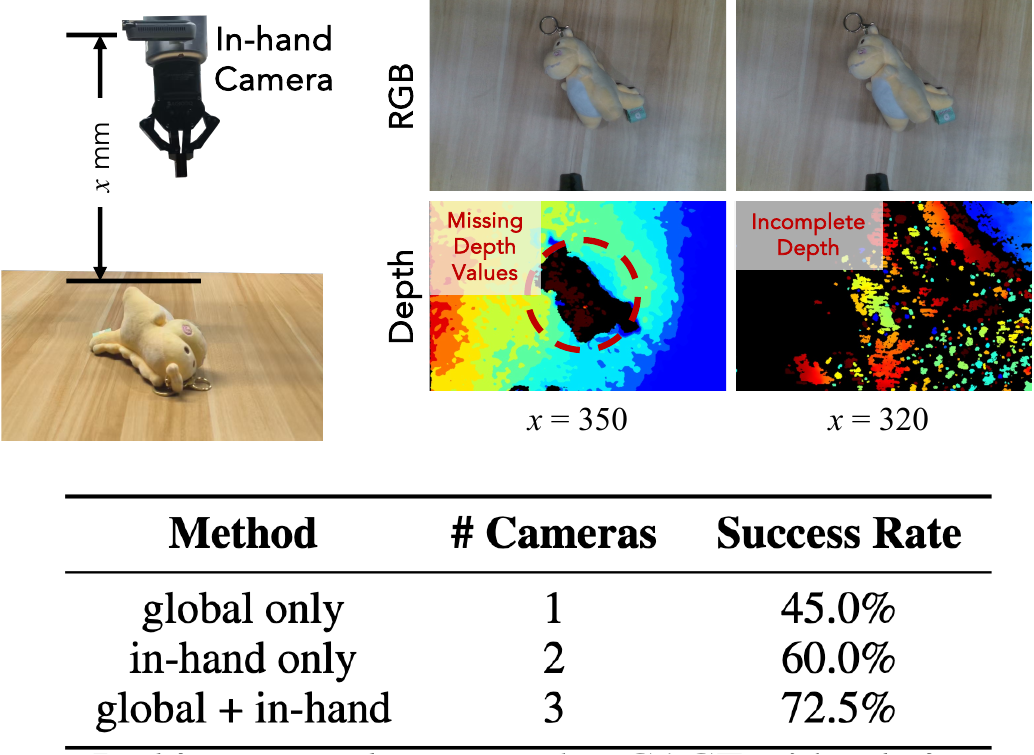}
    \caption{\textbf{Case Study Results}. (Top) In-hand cameras often yield low-quality depth during object interaction, limiting their use for 3D point cloud-based policies. (Bottom) This case study uses CAGE with relative actions, as in the original implementation~\cite{cage}; subsequent experiments use CAGE with absolute actions.}
    \label{fig:inhand}\vspace{-0.6cm}
\end{wrapfigure}
\textbf{In-hand cameras alone are often insufficient for manipulation tasks and may pose additional obstacles on policy learning.} As shown in Fig.~\ref{fig:inhand} (bottom), neither global nor in-hand cameras alone provide adequate observations for 2D image-based policies to achieve strong performance. Recent work~\cite{rise} has demonstrated that using only a global camera enables a 3D imitation policy to outperform 2D multi-view image-based policies, highlighting the importance of 3D information for scene understanding. However, as illustrated in Fig.~\ref{fig:inhand} (top), in-hand cameras may produce incomplete depth information when the robotic arm approaches an object, making them unsuitable for 3D point-cloud-based policies. Consequently, relying solely on in-hand cameras can degrade the performance of the policies, particularly for 3D policies that rely on complete and accurate depth information to achieve superior performance.

\subsection{Additional Experiment Results}\label{app:additional-exp}

\begin{wrapfigure}{r}{0.38\textwidth}
\vspace{-0.7cm}
    \centering
    \includegraphics[width=\linewidth]{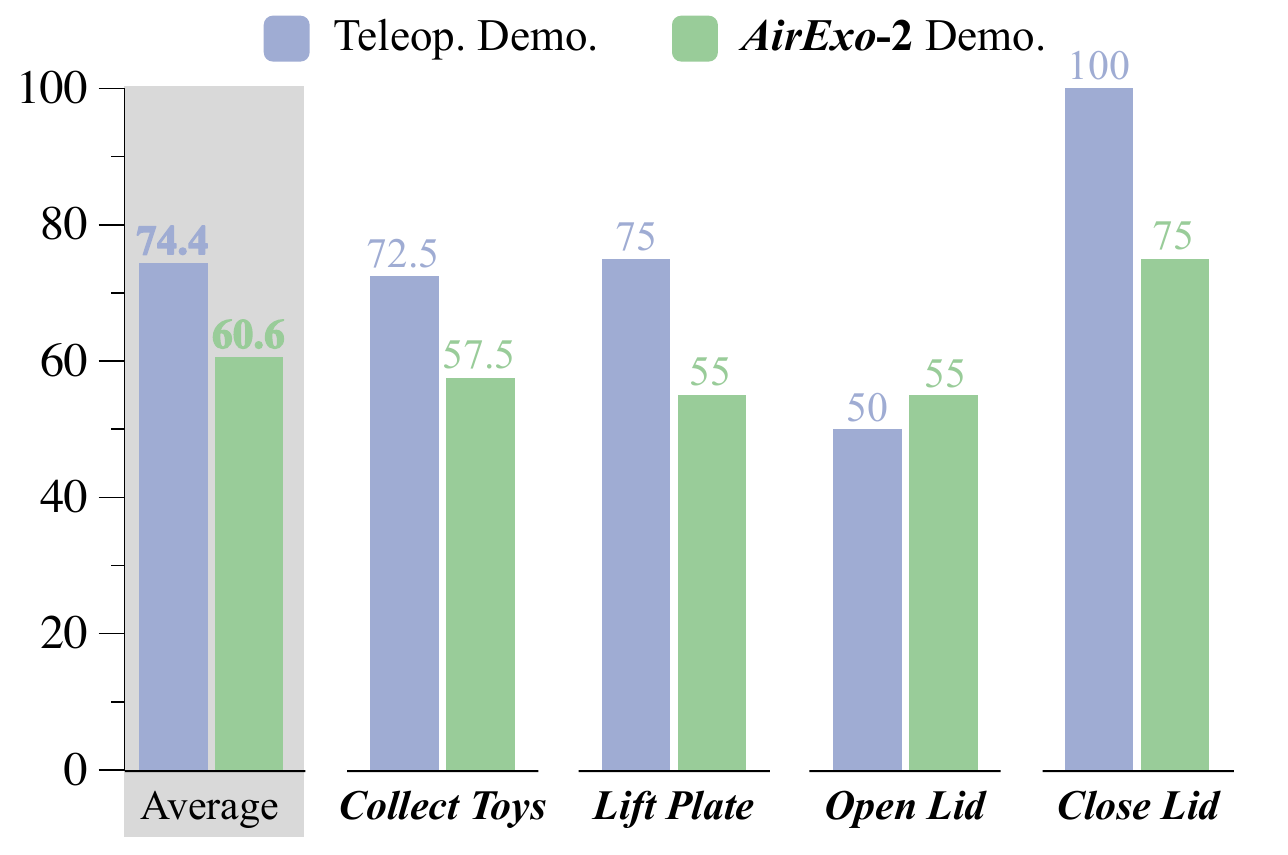}
    \caption{\textbf{RISE Policy Performance with Different Demonstrations}. Generalizable policies are crucial for learning from in-the-wild demonstrations.}
    \label{fig:system-rise}
    \vspace{-0.8cm}
\end{wrapfigure}
\paragraph{Learning from Pseudo-Robot Demonstrations.} We also evaluate the performance of the RISE policy~\cite{rise} using pseudo-robot demonstrations. Specifically, we employ \textbf{\textit{AirExo}-2} to collect an equivalent number of in-the-wild demonstrations, which are then transformed into pseudo-robot data and used to train the RISE policy. The trained policy is deployed on a real robot \textit{without any fine-tuning}. As shown in Fig.~\ref{fig:system-rise} and Tab.~\ref{tab:system}, RISE achieves good zero-shot performance, though with a slightly larger drop compared to \textbf{\textit{RISE}-2}. These findings highlight the critical role of a generalizable policy in successfully transferring manipulation skills from in-the-wild demonstrations to real robots, under the circumstances of no teleoperation data.

\begin{table}[h]
    \centering\footnotesize
    \begin{tabular}{ccc}
        \toprule
        \textbf{Data Source} & \textbf{Policy} & \textbf{Average Success Rate} (4 tasks) $\uparrow$  \\
        \midrule
        \multirow{2}{*}{Teleoperation} & RISE~\cite{rise} & 74.375\% \\
        & \textbf{\textit{RISE}-2} & \textbf{93.750\%} \\
        \midrule
        \multirow{2}{*}{\textbf{\textit{AirExo}-2}} & RISE~\cite{rise} & 60.625\% $^{-13.750\%}$ \\
        & \textbf{\textit{RISE}-2} & \textbf{85.000\%} $^{\text{  }\underline{-8.750\%}}$ \\
        \bottomrule 
    \end{tabular}\vspace{0.1cm}
    \caption{\textbf{Evaluation Results of Different Policies in Learning from Pseudo-Robot Demonstrations}. \textbf{\textit{RISE}-2} shows a smaller performance drop than RISE~\cite{rise} when trained with adapted pseudo-robot demonstrations.}
    \label{tab:system}
    \vspace{-0.4cm}
\end{table}

\paragraph{Action Representations.} We conduct additional experiments on action representations for the \textbf{\textit{Collect Toys}} task, comparing relative and absolute action representations. The results in Tab.~\ref{tab:action-repr} show that while relative action representation sometimes yields better results, absolute action representation provides more stable performance, particularly in terms of generalization. Therefore, we use absolute action representations for Diffusion Policy~\cite{dp} and CAGE~\cite{cage} throughout our experiments, except for the case study in Appendix~\ref{app:case-study}. For $\pi_0$~\cite{pi0}, we still use relative action representation for consistency with its released pre-trained models.

\begin{table}[h]
    \centering\footnotesize
    \begin{tabular}{ccccc}
        \toprule
        \multirow{2}{*}{\textbf{Method}} & \multirow{2}{*}{\textbf{Action Representation}} & \multicolumn{3}{c}{\textbf{Success Rate} $\uparrow$} \\ \cmidrule{3-5}
        & & in-domain & background & object \\
        \midrule
        \multirow{2}{*}{Diffusion Policy~\cite{dp}} & relative & 37.5\% & \textbf{20.0\%} & \textbf{5.0\%}\\
        & absolute & \textbf{40.0\%} & 12.5\% & \textbf{5.0\%}\\ \midrule
        \multirow{2}{*}{CAGE~\cite{cage}} & relative & \textbf{72.5\%} & 32.5\% & 35.0\% \\
        & absolute & 65.0\% & \textbf{45.0\%} & \textbf{42.5\%}\\
        \bottomrule 
    \end{tabular}\vspace{0.1cm}
    \caption{\textbf{Evaluation Results of Different Action Representations on the \textit{Collect Toys} Task.} Generally, absolute action representation leads to a more stable performance than relative action representation.}
    \label{tab:action-repr}
    \vspace{-0.4cm}
\end{table}

\subsection{Failure Analysis}

\paragraph{Imprecise Action.} During the policy rollouts, we observe that most failures stem from imprecise predicted actions. For example, in the \textbf{\textit{Open Lid}} task, the lid protrudes only about 1.5 cm from the coffee machine body, requiring highly accurate end-effector control from the visuomotor policies. At the same time, the left gripper must stabilize the machine to prevent it from shifting on the table while the right arm applies force to lift the lid. Baseline failures typically arise from inaccuracy in either arm --- where the left arm fails to stabilize the coffee machine or the right arm fails to open the lid. In contrast, our \textbf{\textit{RISE}-2} policy produces more precise actions, resulting in fewer failures, even when trained with adapted in-the-wild data.

\paragraph{Risk of Visual Overfitting to Robot States.} We also observe that in bimanual manipulation tasks, since both arms occupy a large portion of the observation image, \textit{there is an increased risk of visuomotor policies overfitting to robot states}. This occurs even without proprioceptive inputs, as the policy may implicitly infer actions from the visual configuration of the arms. The issue is further compounded by limited training data: with only 50 demonstrations, most baselines fail to learn sufficiently accurate actions, leading to low phase success rates. Such overfitting also reduces the likelihood of performing recovery behaviors after phase failures, which becomes especially evident in long-horizon and complex tasks such as \textbf{\textit{Serve Steak}}, ultimately causing frequent task failures. In contrast, \textbf{\textit{RISE}-2} produces more precise actions, resulting in higher success rates. Nevertheless, all policies still exhibit varying degrees of overfitting to robot visual state and struggle to recover from failures in the absence of recovery demonstrations, highlighting an important direction for developing more generalizable robotic visuomotor policies.

\section{Author Contributions}

The author contributions for this paper are as follows. For each item, authors are listed in order of contribution, and \underline{underlined names} indicate equal contributions.

\begin{itemize}
    \item \textbf{Project Lead}: Hongjie Fang.
    \item \textbf{System Design and Implementation}.
    \begin{itemize}
    \item \textbf{\textit{AirExo}-2 Hardware}: Yiming Wang, Hongjie Fang, Hao-Shu Fang.
    \item \textbf{\textit{AirExo}-2 Visual Adaptors}: Jingjing Chen, Hongjie Fang.
    \item \textbf{\textit{RISE}-2 Policy}: Chenxi Wang, Hongjie Fang, Jingjing Chen, Shangning Xia.
    \end{itemize}
    \item \textbf{Experiments}:
    \begin{itemize}
        \item \textbf{Design}: \underline{Hongjie Fang}, \underline{Chenxi Wang}, \underline{Jingjing Chen}, Yiming Wang, Hao-Shu Fang.
        \item \textbf{Platform Setup}: Hongjie Fang, Yiming Wang.
        \item \textbf{Data Collection}: \underline{Hongjie Fang}, \underline{Chenxi Wang}, \underline{Jingjing Chen}.
        \item \textbf{Data Processing}: Jingjing Chen, Zihao He, Xiyan Yi, Yunhan Guo.
        \item \textbf{Baselines}: \underline{Hongjie Fang}, \underline{Shangning Xia}, Chenxi Wang, Jingjing Chen.
        \item \textbf{Policy Training}: \underline{Hongjie Fang}, \underline{Chenxi Wang}, \underline{Shangning Xia}, Jingjing Chen, Xinyu Zhan, Lixin Yang.
        \item \textbf{Policy Evaluation}: \underline{Hongjie Fang}, \underline{Chenxi Wang}, \underline{Jingjing Chen}.
        \item \textbf{User Study}: Hongjie Fang, Chenxi Wang, Jingjing Chen, Xiyan Yi.
        \item \textbf{Accuracy Analysis}: Hongjie Fang, Jun Lv, Jingjing Chen, Yiming Wang.
        \item \textbf{Scalability Analysis}: \underline{Hongjie Fang}, \underline{Jingjing Chen}, Chenxi Wang, Hao-Shu Fang.
    \end{itemize}
    \item \textbf{Writing}: Hongjie Fang, Chenxi Wang, Jingjing Chen, Yiming Wang, Hao-Shu Fang.
    \item \textbf{Project Advisor}: \underline{Hao-Shu Fang}, \underline{Cewu Lu}, Weiming Wang.
\end{itemize}

\end{document}